\pdfoutput=1

\documentclass[11pt]{article}

\usepackage[final]{acl}

\usepackage{times}
\usepackage{latexsym}
\usepackage{multirow}
\usepackage[normalem]{ulem}
\useunder{\uline}{\ul}{}
\usepackage{makecell}
\usepackage{microtype}
\usepackage{geometry}
\usepackage[T1]{fontenc}

\usepackage[utf8]{inputenc}

\usepackage{microtype}

\usepackage{inconsolata}

\usepackage{graphicx}
\usepackage{multirow}
\usepackage[normalem]{ulem}
\useunder{\uline}{\ul}{}
\usepackage{booktabs}
\usepackage{colortbl}
\usepackage{amsmath}
\usepackage{amssymb,amsthm}
\usepackage{authblk}  
\usepackage{xspace}    
\usepackage{ragged2e}  
\usepackage{setspace}  
\usepackage{footmisc}
\usepackage{etoolbox}

\makeatletter
\patchcmd{\@makefntext}{\indent}{\noindent}{}{}  
\makeatother

\newcommand*{\affmark}[1][*]{\textsuperscript{#1}} 
\newcommand*{\email}[1]{\texttt{#1}} 

\title{See-Saw Modality Balance: See Gradient, and Sew Impaired Vision-Language Balance to Mitigate Dominant Modality Bias}

\author{\begin{large}
  \textbf{JuneHyoung Kwon\affmark[1]\textsuperscript{*}},  
  \textbf{MiHyeon Kim\affmark[1]\textsuperscript{*}\textsuperscript{†}},  
  \textbf{Eunju Lee\affmark[2]},  
  \textbf{Juhwan Choi\affmark[1]},  
  \textbf{YoungBin Kim\affmark[1,2]}  
  \end{large} \\
}
\vspace{-1mm}
\begin{normalsize}
\affil{\textsuperscript{1} Department of Artificial Intelligence, Chung-Ang University}
\affil{\textsuperscript{2} Graduate School of Advanced Imaging Sciences, Multimedia and Film, Chung-Ang University}
\affil{\email{\{dirchdmltnv, mh10967, dmswn5829, gold5230, ybkim85\}@cau.ac.kr}}
\end{normalsize}
\date{}

\begin{document}

\maketitle


\footnotetext{\textsuperscript{*}Equal contribution.}
\footnotetext{\textsuperscript{†}Currently at: KT CORPORATION, \email{mihyeon.gim@kt.com}.}

\begin{abstract}
Vision-language (VL) models have demonstrated strong performance across various tasks. However, these models often rely on a specific modality for predictions, leading to ``dominant modality bias.'' This bias significantly hurts performance, especially when one modality is impaired. In this study, we analyze model behavior under dominant modality bias and theoretically show that unaligned gradients or differences in gradient \textit{magnitudes} prevent balanced convergence of the loss. Based on these findings, we propose a novel framework, \textbf{\textsc{BalGrad}} to mitigate dominant modality bias. Our approach includes inter-modality gradient reweighting, adjusting the gradient of KL divergence based on each modality's contribution, and inter-task gradient projection to align task \textit{directions} in a non-conflicting manner. Experiments on UPMC Food-101, Hateful Memes, and MM-IMDb datasets confirm that \textsc{BalGrad} effectively alleviates over-reliance on specific modalities when making predictions.
\end{abstract}

\section{Introduction}

Vision-language (VL) models combine image and text modalities, resulting in powerful multi-modal representations. Owing to this integration of two modalities, these models can achieve higher performance in vision-language tasks. Recently, leveraging extensive datasets, VL models have demonstrated remarkable performance across various tasks such as image captioning~\cite{hu2022scaling}, visual question answering~\cite{khademi2023mm}, and cross-modal retrieval~\cite{liu2022universal}, showcasing their capability to harness the complementary strengths of visual and textual data.

\begin{figure}[t]
  \includegraphics[width=\columnwidth]{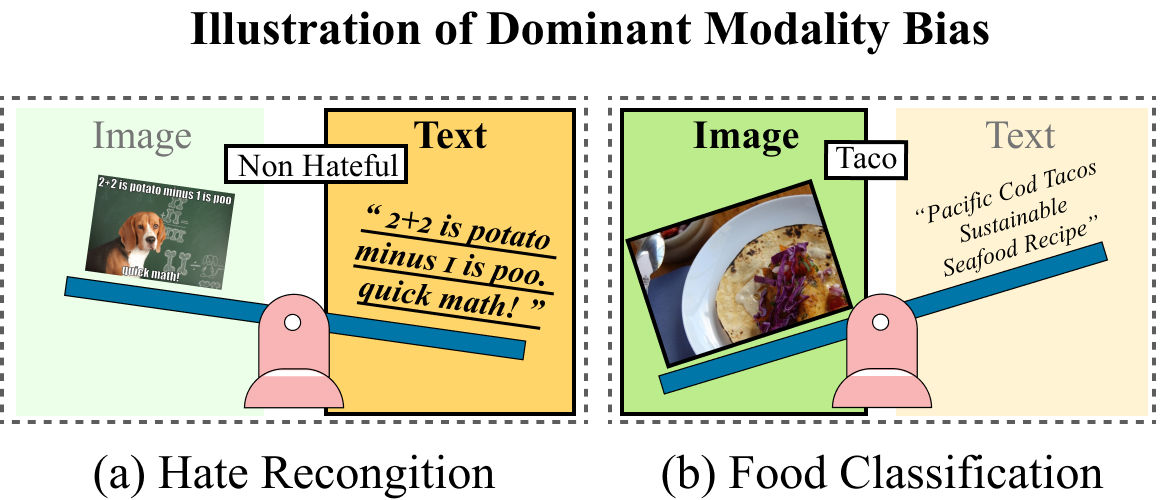}
  \caption{Conceptual visualization of dominant modality bias. The key modality differs by task: \textbf{(a)} For the hate recognition task, text descriptions of memes lead, while \textbf{(b)} for the food classification task, food images play a crucial role in prediction.}
  \label{fig:concept}
\end{figure}

However, these models often rely on a single modality rather than treating and utilizing them equally, leading to the dominance of a certain modality on the overall performance. A conceptual overview of this effect can be seen in Figure~\ref{fig:concept}. This phenomenon, where a specific modality disproportionately influences the model's outcomes, is referred to as ``dominant modality bias’’~\cite{woo2023towards}. For instance, VL models tend to be biased towards the text modality when recognizing hate expressions~\cite{kiela2020hateful, aggarwal2024text}, thereby limiting the VL model's ability to effectively integrate and interpret images.

This bias behaves particularly detrimentally when one modality is impaired, such as when data is noisy and it is difficult to gather paired data~\cite{garg2022multimodality, woo2023towards, yang2023quantifying}. This issue is common in real-world scenarios due to privacy-related data sharing restrictions or stringent data storage policies~\cite{voigt2017eu} and can severely degrade the model's performance. Additionally, the failure to sufficiently explore the weak modality limits the overall performance of the VL model~\cite{wang2020makes, huang2022modality, peng2022balanced}, highlighting the need for robust solutions to mitigate dominant modality bias.

To address this issue and balance the information between modalities, numerous studies have been conducted. Several studies have focused on modulating the gradients of each encoder based on the confidence of individual modalities~\cite{peng2022balanced, li2023boosting}. Other approaches have involved training multimodal models using the best-performing learning rates from unimodal models~\cite{yao2022modality}. However, these methods often induce negative transfer~\cite{wang2019characterizing,yu2020gradient}, which occurs when the model's performance decreases with the addition of modality data compared to solely using unimodal data. 


We first analyze the behavior of models after the dominant modality bias has taken root. Our analysis reveals that certain modalities are more crucial to target performance and observes that the dominant and weak modalities converge at different rates during training. Additionally, we theoretically demonstrate that the balanced convergence of the loss is influenced by both the \textit{magnitude} and \textit{direction} of the gradient. Based on these findings, we propose \textbf{\textsc{BalGrad} (Balancing Gradients)} to mitigate dominant modality bias. Firstly, we adopt a mutual KL divergence between the predictions of each modality to ensure balanced updates. However, a naive approach that equally aligns the distributions of two modalities can hinder the representation learning of each modality. To address this, we introduce \textbf{inter-modality gradient reweighting}, which adjusts the \textit{magnitude} of the gradient of the KL divergence term based on the learning status of each modality. Additionally, we propose \textbf{inter-task gradient projection}, which updates the gradient of the target task to establish a balance between both modalities. We project the target task's gradient in a \textit{direction} orthogonal to the KL divergence gradient if a conflict between the gradients occurs, encouraging stabilized training between the two modalities.

We evaluate the effectiveness of \textsc{BalGrad} on models using three vision-language datasets: UPMC Food-101~\cite{wang2015recipe}, Hateful Memes~\cite{kiela2020hateful}, and MM-IMDb~\cite{arevalo2017gated}. To simulate the influence of individual modalities, we conduct experiments under conditions where specific modalities are missing or impaired by noise. The experimental results demonstrate that the proposed method reduces the gap between the modalities while avoiding negative transfer. The contributions of our proposed method are as follows:

\begin{itemize}
\item We analyze the dominant modality bias and theoretically demonstrate that the balanced convergence of loss is influenced by both the \textit{magnitude} and \textit{direction} of the gradient.
\item We propose \textsc{BalGrad}, which reweights the gradients between modalities to ensure stable convergence and projects the target task's gradient to avoid conflicts that hinder balanced learning.
\item Experimental results across UPMC Food-101, Hateful Memes, and MM-IMDb under different impaired conditions confirm the effectiveness of our proposed method in mitigating dominant modality bias.
\end{itemize}

\section{Related Work}

\begin{figure*}[t]
  \includegraphics[width=\textwidth]{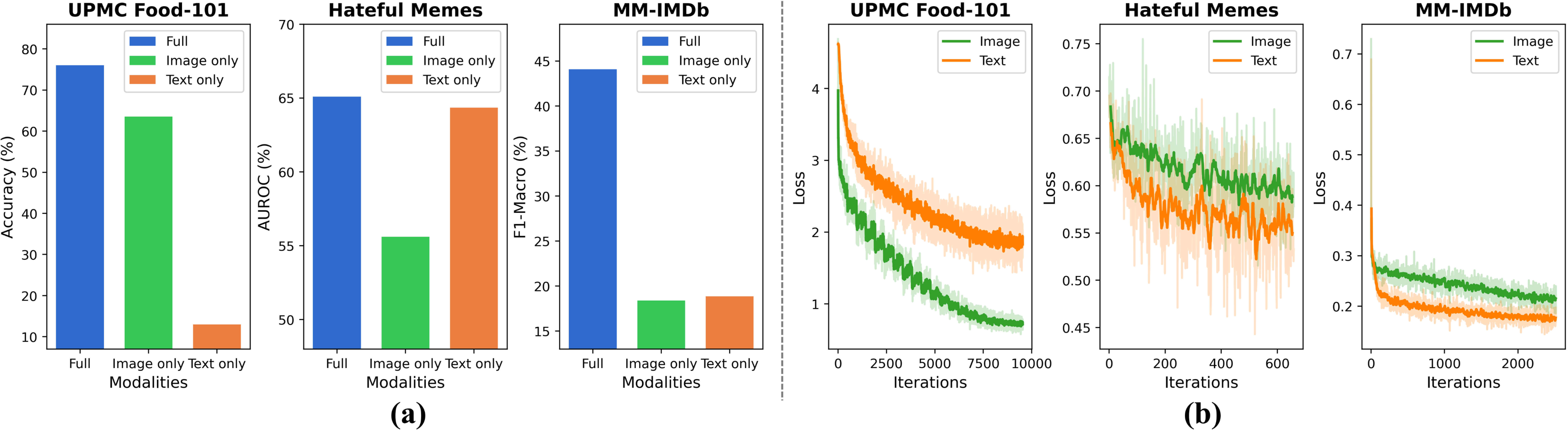}
  \caption{Experimental results on the UPMC Food-101, Hateful Memes, and MM-IMDb datasets in the presence of dominant modality bias. \textbf{(a)} Performance visualization under different missing conditions (full, image only (missing text), text only (missing image)) for each dataset. \textbf{(b)} Illustration of learning curves for each modality across datasets.}
  \label{fig:modal_gap}
\end{figure*}

In multimodal models, such as VL models, a bias towards a preferred or easier-to-learn modality often leads to the under-exploration of others~\cite{wang2020makes, huang2022modality, peng2022balanced}. Studies have analyzed this, noting that multimodal models are prone to overfitting and show discrepancies in generalization across modalities~\cite{wang2020makes}. Differences in convergence speeds also contribute to this bias~\cite{yao2022modality, wu2022characterizing}. An early study in this field finds that certain modalities, correlating with their network’s random initialization, dominate the learning process~\cite{huang2022modality}, while other researchers attribute the preference to unimodal representation margins and insufficient integration of modalities~\cite{yang2023quantifying}. Another line of study highlights that spurious correlations with instance labels cause imbalances in modality utilization~\cite{guo2023modality}. In this paper, we identify that the dominant modality bias in VL models arises from the influence of gradient magnitude and direction on the model's loss function, hindering balanced learning across modalities.

In response to the challenge of balancing modalities in multimodal learning, various strategies have been proposed. MSLR suggests using different optimal learning rates for each modality during multimodal learning to enhance performance~\cite{yao2022modality}. Another approach involves using a conditional utilization rate to re-scale modality features, ensuring balanced contributions from each modality~\cite{wu2022characterizing}. Gradient blending optimizes the mixing of modalities based on the model's overfitting behavior~\cite{wang2020makes}. OGM-GE adaptively controls the optimization process using modality-specific confidence scores~\cite{peng2022balanced}. AGM employs Shapley values to modulate gradients through mono-modal responses, aiming to balance the learning process across modalities~\cite{li2023boosting}. However, these methods often lack consideration of negative transfer and may introduce adverse effects. In this paper, we propose \textsc{BalGrad}, which reweights gradients considering the learning status of each modality and projects the gradients to mitigate dominant modality bias without disrupting the balance between modalities.

\section{Method}
In this section, we analyze the dominant modality bias and propose \textsc{BalGrad} to mitigate such bias. In Section~\ref{analysis}, we observe the behavior of VL models and theoretically demonstrate the factors influencing balanced loss convergence. In Section~\ref{balgrad}, based on these findings, we introduce \textsc{BalGrad}, which reweights and projects gradients to ensure balanced learning across modalities.

\subsection{Analysis of Dominant Modality Bias}
\label{analysis}

We introduce a controlled experiment to analyze the behavior of VL models biased by dominant modality. We denote the training dataset as $\mathcal{D} = \{(x_i, y_i)\}_{i=1}^N$, where $x_i = (x_i^v, x_i^l)$ is a pair of data from the image and text modalities, respectively, and $y_i$ represents the label. We extract features from the image and text encoders, passing them through their respective embedding layers, $h_v(\cdot)$ and $h_l(\cdot)$. These embeddings are then fused via concatenation and passed through a classifier, $f_{\mathcal{T}}(\cdot)$, to yield the predicted probability $p_{\mathcal{T}}$. Details on the architecture and training scheme are provided in the Appendix \ref{implementation details}. 

\noindent \textbf{Analysis on Performance Gap.} To analyze the impact of individual modalities on the performance of VL models, we mute one modality by inputting empty values at the data level, rendering it non-informative. This method is applied while testing on the UPMC Food-101, Hateful Memes, and MM-IMDb datasets. The experimental results in Figure~\ref{fig:modal_gap} (a) show a significant performance drop when a specific modality is missing. In the case of UPMC Food-101, the image modality significantly influences the overall performance, while in Hateful Memes, the text modality plays a more crucial role. Conversely, the performance drop is relatively minor when the weak modality (\textit{text} for \textit{UPMC Food-101} and \textit{image} for \textit{Hateful Memes}) is missing. In contrast, for MM-IMDb, the performance drop is similar when either modality is missing, indicating that the model is not biased towards a specific modality.

\noindent \textbf{Analysis on Training Dynamics.} To observe the loss dynamics of each modality during the training phase, we add linear classifiers $f_v(\cdot)$ and $f_l(\cdot)$ on top of the image and text embedding layers, respectively. These classifiers output probabilities $p_i^v$ and $p_i^l$, which are then used to predict the label $y_i$, and each target objective is represented as $\mathcal{L}^v_{\mathcal{T}}$ and $\mathcal{L}^l_{\mathcal{T}}$, respectively. We find that the loss of the dominant modality decreases rapidly, while the loss of the weak modality decreases relatively slowly, as shown in Figure~\ref{fig:modal_gap} (b). For MM-IMDb specifically, the loss gap decreases as training iterations increase, demonstrating that the model is not biased toward any single modality. This indicates that, during training, one modality is overly exploited while the other modality is relatively underexplored, which is consistent with previous research~\cite{wang2020makes, huang2022modality, peng2022balanced}. We conjecture that this phenomenon appears inherently task-dependent, with the VL model inclined to update based on the easy-to-learn modality that can quickly reduce the loss~\cite{arpit2017closer, nam2020learning}.

\begin{figure*}[t]
  \includegraphics[width=\textwidth]{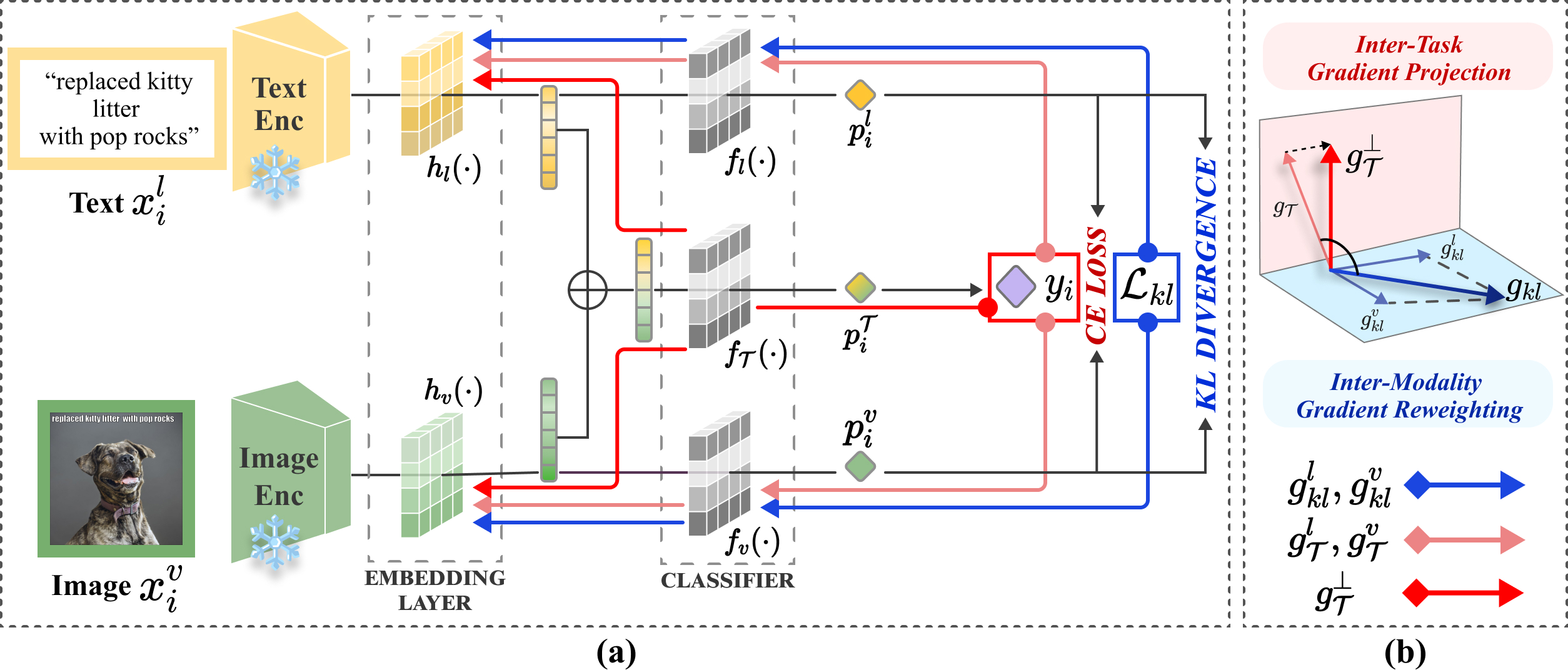}
  \caption{\textbf{(a)} The overall training framework of our proposed \textsc{BalGrad}. The final classifier $f_{\mathcal{T}}(\cdot)$ is updated with the gradient $g^{\perp}_{\mathcal{T}}$ for cross entropy (CE) loss. The image and text embedding layers $h_v(\cdot), h_l(\cdot)$ are also updated with $g^{\perp}_{\mathcal{T}}$ along with the gradients of the CE loss for each modality $g^v_{\mathcal{T}}, g^l_{\mathcal{T}}$, and the gradients of the KL divergence between the two modalities' predictions $g^v_{kl}, g^l_{kl}$. \textbf{(b)} Inter-modality gradient reweighting adjusts the magnitudes of $g^v_{kl}$ and $g^l_{kl}$ to obtain $g_{kl}$. If a conflict occurs, we project $g^{\perp}_{\mathcal{T}}$ on the orthogonal direction of $g_{kl}$ by inter-task gradient projection.}
  \label{fig:overall}
\end{figure*}

\noindent \textbf{Theoretical Analysis of Gradient Influence.}
To theoretically analyze why VL models struggle to balance the utilization of both modalities, we examine the loss reduction in terms of gradient updates. The loss function for a target is defined as $\mathcal{L}(\theta_v, \theta_l, \theta_{\mathcal{T}})$, where $\theta_v$ and $\theta_l$ are the parameters of image and text embedding layers, respectively, and the $\theta_{\mathcal{T}}$ represents the parameters of the classifier $f_{\mathcal{T}}(\cdot)$. The objective is to find the optimal parameters $\Theta=\{\theta_v, \theta_l, \theta_{\mathcal{T}}\}$ that minimize $\mathcal{L}(\theta_v, \theta_l, \theta_{\mathcal{T}})$. To analyze how each modality contributes to the overall loss reduction, we decompose the target task loss gradient with respect to the model parameters $\Theta$ into modality-specific components, denoted by $\mathcal{G^\tau}=\{g_l, g_v, g_{\mathcal{T}}\}$. These partial gradients capture the influence of linguistic, visual, and task-related parameters, respectively, under standard gradient-descent updates. Additionally, $g_{\mathcal{T}} = \sum_{i \in \{v, l\}} \triangledown_{\theta_i} \hat{y} \triangledown_{\hat{y}} p_{\mathcal{T}} \triangledown_{p_{\mathcal{T}}} \mathcal{L} = \sum_{i \in \{v, l\}} g_{\mathcal{T}}^i$ denotes the gradient for parameters $\theta_{\mathcal{T}}$ of the linear classifier $f_\mathcal{T}(\cdot)$, where $g_{\mathcal{T}}^i$ denotes the gradient of each modality in $f_\mathcal{T}(\cdot)$. We theoretically analyze how the target objective is influenced by the varying magnitudes and directions of gradients for each modality.

\noindent \textbf{Proposition 1. (Gradient Effect on Change of Loss)} 
Let the parameters $\theta_v, \theta_l,$ and $\theta_{\mathcal{T}}$ of a multimodal model be updated with gradients $g_v, g_l,$ and $g_{\mathcal{T}}$ using a sufficiently small step size $\lambda>0$, resulting in updated parameters $\hat{\theta}_v, \hat{\theta}_l,$ and $\hat{\theta}_{\mathcal{T}}$. Then the change in the loss function satisfies
\begin{equation}
\begin{aligned}
    \triangle\mathcal{L}
    \;&=\;-2 \,\lambda \,\bigl(g_{\mathcal{T}}^v \cdot g_{\mathcal{T}}^l\bigr)
    \\&
    -\,\lambda \sum_{\,i\in\{v,l,\mathcal{T}\}}\!
      \Bigl(\,g_i \cdot g_i \;+\; g_{\mathcal{T}}^i \cdot g_{\mathcal{T}}^i\Bigr)
    \;+\;
    O(\lambda^2),
\end{aligned}
\end{equation}
where the cross term $-2\,\lambda\,(g_{\mathcal{T}}^v)\cdot(g_{\mathcal{T}}^l)$ captures the interaction between the visual and language gradients and the magnitudes and directions of each gradient $g_{\mathcal{T}}^v$ and $g_{\mathcal{T}}^l$ governs how much the overall loss is reduced.
\begin{proof}
See Appendix ~\ref{proof1}
\end{proof}

If the gradients for the two modalities $g_{\mathcal{T}}^v$ and $g_{\mathcal{T}}^l$ do not align well, meaning they have conflicting directions or have significantly different magnitudes, the loss reduction will not be balanced. Gradients with larger magnitudes substantially impact loss reduction, while gradients with directions that align more closely between modalities facilitate more effective joint learning. Consequently, the loss is likely to decrease more under the influence of the dominant modality, leading to an uneven contribution from each modality.

\subsection{\textsc{BalGrad}}
\label{balgrad}

Based on the findings above, we propose \textsc{BalGrad} to mitigate the dominant modality bias, which consists of two components: inter-modality gradient reweighting and inter-task gradient projection. Inter-modality gradient reweighting addresses the imbalance caused by different gradient \textit{magnitudes}, ensuring more equal contributions from each modality. Inter-task gradient projection aligns the gradient \textit{directions} of the modalities, facilitating more effective joint learning and preventing the dominant modality from disproportionately influencing loss reduction. The overall process of \textsc{BalGrad} can be seen in Figure~\ref{fig:overall}. 

\subsubsection{Inter-modality Gradient Reweighting}

Standard VL models lack the consideration to ensure that both modalities are updated equally, leading to the stronger modality dominating the training phase, as we observed in the previous section. Therefore, inspired by knowledge distillation~\cite{hinton2015distilling, zhang2018deep, phuong2019towards}, we aim to balance the gradients received from each modality by aligning the distributions of their predictions. To achieve this, we compute the mutual Kullback-Leibler (KL) divergence between the predictions $p^v_i$ and $p^l_i$ of the two modalities. This involves aligning the predictions of the image modality with those of the text modality and vice versa. The KL divergence from $p^l_i$ to $p^v_i$ is as follows:

\begin{equation}
\mathcal{L}^l_{kl}=-\sum_{i}p^l_i log \frac{p^v_i}{p^l_i}
\end{equation}

We also compute $\mathcal{L}^v_{kl}$ in the same manner. We represent the gradients of $\mathcal{L}^l_{kl}$ and $\mathcal{L}^v_{kl}$ as $g^l_{kl} = \nabla \mathcal{L}^l_{kl}$ and $g^v_{kl} = \nabla \mathcal{L}^v_{kl}$, respectively. In this way, each modality's embedding layer learns to correctly predict the label and match the probability estimate of other modalities, thereby alleviating the severe imbalance. However, symmetrically aligning the distributions between the two modalities overlooks the differences in their convergence status, as observed in Section~\ref{analysis}. This can cause the layers of the faster-converging modality to be hindered in their representation learning, leading to performance degradation. Therefore, we propose an \textbf{inter-modality gradient reweighting} method that adjusts the magnitude to which each modality receives the KL divergence gradient based on its contribution to the learning objective. We reweight the gradient of the KL divergence term for $p^l_i$ to $p^v_i$ and $p^v_i$ to $p^l_i$ using the following terms, respectively:

\begin{equation}
\mathcal{W}^l= \frac{\mathcal{L}^l_{\mathcal{T}}} {\mathcal{L}^v_{\mathcal{T}} + \mathcal{L}^l_{\mathcal{T}}}, \ \mathcal{W}^v= \frac{\mathcal{L}^v_{\mathcal{T}}} {\mathcal{L}^v_{\mathcal{T}} + \mathcal{L}^l_{\mathcal{T}}}
\end{equation}

In this configuration, if the target task loss for a modality is low (i.e., it has converged more), the gradient receives a lower weight. This ensures that the gradient of the weak modality is updated more toward matching the dominant modality's prediction, thereby reducing the training gap. In contrast, the dominant modality receives less influence from the underperforming predictions, allowing it to effectively learn its representation. 
Additionally, to ensure that each modality is trained for the target task independently, we introduce an additional term that increases the reweighting factor as iteration $t$ progresses. This ensures that the impact of mutual learning grows over time, allowing individual encoders to learn effectively in the initial stages and progressively encouraging balanced learning between modalities. The final reweighted gradient for the KL divergence is as follows:

\begin{equation}
g_{kl} = (\gamma + \frac {\gamma} {1 + e^{-t}})(\mathcal{W}^l g^l_{kl} + \mathcal{W}^v g^v_{kl})
\end{equation}
$\gamma$ is the initial weighting factor, and we set $\gamma=1/2$.

\subsubsection{Inter-task Gradient Projection}

Proposition 1 highlights that properly aligning gradients with different directions and magnitudes is crucial for effective joint learning. However, when gradients are not aligned and exhibit negative cosine similarity, known as conflicting gradients, the optimization process becomes suboptimal~\cite{yu2020gradient,shi2023recon}. Such conflicts can arise between the gradients of different tasks, potentially causing the dominant gradient to overwhelm the optimization process at the expense of the other task's performance.

For our case, as confirmed in Section~\ref{analysis}, the target task, which is $\mathcal{L}_{\mathcal{T}}$ alone, fails to balance the modalities and fully explore the weak modality. Therefore, we introduce the balance between the predictions of each modality as an additional task. However, as mentioned earlier, naive joint training can cause conflict between the gradient of the target task and KL divergence (i.e., $g_{\mathcal{T}}$ and $g_{kl}$).\\

\noindent\textbf{Proposition 2. (Gradient Conflicts on Loss Reduction with KL Loss)}
Let $\mathcal{G}^\tau = \{\,g_v,\,g_l,\,g_\tau\}$ and $\mathcal{G}^{kl} = \{\,g_v^{kl},\,g_l^{kl},\,0\}$ be the gradients from a target loss $\mathcal{L}_\tau$ and a KL loss $\mathcal{L}_{kl}$, respectively, with parameters $\theta = [\theta_v,\theta_l,\theta_{\tau}]^\top$. Assume the parameters are updated by gradient descent with a small step size $\lambda>0$: $\theta_v' = \theta_v - \lambda\,\bigl(g_v + g_v^{kl}\bigr),
\theta_l' = \theta_l - \lambda\,\bigl(g_l + g_l^{kl}\bigr),
\theta_{\tau}' = \theta_{\tau} - \lambda\,g_{\tau}$.
Then, for the combined loss $\mathcal{L} = \mathcal{L}_\tau + \mathcal{L}_{kl}$, the change in the loss is
\begin{equation}
\begin{aligned}
    \Delta \mathcal{L} &=\mathcal{L}\bigl(\theta'\bigr) -\mathcal{L}\bigl(\theta\bigr) \\&=-\,\lambda \Bigl(\|\mathcal{G}^\tau\|^2 +\,\|\mathcal{G}^{kl}\|^2 +\,2\,(\mathcal{G}^\tau)^\top \mathcal{G}^{kl}\Bigr)\\&+ O(\lambda^2).
\end{aligned}
\end{equation}
In particular, if $(\mathcal{G}^\tau)^\top \mathcal{G}^{kl} < 0,$ the gradients from the target and KL losses \emph{conflict}, reducing the effective loss reduction.
\begin{proof}
See Appendix ~\ref{proof2}
\end{proof} 

Building upon Proposition 2, we aim to ensure that the gradient of the target task does not disrupt the balance between modalities. Specifically, we propose \textbf{inter-task gradient projection}, which projects $g_{\mathcal{T}}$ onto $g_{kl}$ in a non-conflicting manner. First, we consider the relationship between the two gradients to determine if they conflict and compute the cosine similarity between the two gradients. If $g_{\mathcal{T}} \cdot g_{kl} \ge 0$, we assume that $g_{\mathcal{T}}$ is being updated in a direction that aligns with modality balance, and we use the original $g_{\mathcal{T}}$ for updating the model. Conversely, if $g_{\mathcal{T}} \cdot g_{kl} < 0$, indicating a potential disruption to the balance between modalities, we project $g_{\mathcal{T}}$ in a direction orthogonal to $g_{kl}$. This process can be represented as follows: 

\begin{equation}
g_{\mathcal{T}}^\perp =
\begin{cases}
g_{\mathcal{T}} - \left( \frac{g_{\mathcal{T}} \cdot g_{kl}}{\|g_{kl}\|^2} \right) g_{kl}, & \mbox{if }g_{\mathcal{T}} \cdot g_{kl} <0 \\
g_{\mathcal{T}}, & \mbox{otherwise}
\end{cases}
\end{equation}

This projection ensures that $g_{\mathcal{T}}^\perp$ is adjusted to maintain the balance between the modalities while preventing conflicts with $g_{kl}$. In a nutshell, the proposed \textsc{BalGrad} allows for extensively learning different modalities and tasks, effectively optimizing the target task while maintaining the balance between the modalities.

\section{Experiments}

\subsection{Experimental Setup}

We conduct experiments on three vision-language datasets: UPMC Food-101~\cite{wang2015recipe}, Hateful Memes~\cite{kiela2020hateful}, and MM-IMDb~\cite{arevalo2017gated}. For image and text encoding, we utilize ViT~\cite{dosovitskiy2020image} and BERT~\cite{devlin2018bert}, respectively, employing a late concatenation architecture for final predictions. To minimize extensive fine-tuning, we adopt linear probing, freezing all encoder parameters and training only the embedding and classifier layers. Further implementation details are provided in Appendix \ref{implementation details}.

To assess the robustness of the VL model against dominant modality bias, we introduced two impaired conditions: missing and noisy. For the missing modality, empty strings were used for text and zero pixels for images~\cite{lee2023multimodal}. In the noisy condition, 30\% salt and pepper noise is added to images~\cite{lim2023biasadv}, and 15\% of text tokens were randomly deleted~\cite{manolache2021date,yuan2022hype}. All experiments were conducted with the model trained on unimpaired full modality data, with impairments applied to the entire data of a specific modality during testing.  Further implementation details are provided in Appendix \ref{additional experimental results}.

\subsection{Experimental Results}

\begin{table*}[t]
\centering
\resizebox{\textwidth}{!}{%
\addtolength{\tabcolsep}{-0.4em}
\renewcommand{\arraystretch}{1.1}
\begin{tabular}{cc|ccccc|ccccc|ccccc}
\toprule
\rowcolor[HTML]{FFFFFF} 
\multicolumn{2}{c|}{\cellcolor[HTML]{FFFFFF}}                                    & \multicolumn{5}{c|}{\cellcolor[HTML]{FFFFFF}\textbf{UPMC Food-101}}                    & \multicolumn{5}{c|}{\cellcolor[HTML]{FFFFFF}\textbf{Hateful Memes}}              & \multicolumn{5}{c}{\cellcolor[HTML]{FFFFFF}\textbf{MM-IMDb}}                       \\ \cline{3-17} 
\rowcolor[HTML]{FFFFFF} 
\multicolumn{2}{c|}{\multirow{-2}{*}{\cellcolor[HTML]{FFFFFF}Modality}} & Baseline    & MSLR        & OGM-GE & AGM            & \textbf{\textsc{BalGrad}}        & Baseline   & MSLR        & OGM-GE          & AGM          & \textbf{\textsc{BalGrad}}        & Baseline       & MSLR           & OGM-GE         & AGM            & \textbf{\textsc{BalGrad}}  \\ 
\midrule
\rowcolor[HTML]{FFFFFF} 
\multicolumn{2}{c|}{\cellcolor[HTML]{FFFFFF}Full}                                & 76.01       & 78.43       & 77.42  & {\ul 78.93}          & \textbf{80.32}       & 65.10      & 65.58       & {\ul 66.70}     & 64.69        & \textbf{67.35}       & \textbf{44.09}          & \textbf{44.09}          & 42.22          & 43.93 & 43.19          \\
\midrule
\rowcolor[HTML]{FFFFFF} 
\cellcolor[HTML]{FFFFFF}                                    & Image              & 12.99 & 20.52       & 13.86  & {\ul 22.60}          & \textbf{25.49}       & 64.34      & 66.04       & \textcolor{gray}{\textbf{66.83*}} & \textcolor{gray}{\ul 66.25*} & 65.86                & 18.85          & {\ul 19.26}    & \textbf{24.48} & 17.57          & 18.81          \\
\rowcolor[HTML]{FFFFFF} 
\cellcolor[HTML]{FFFFFF}                                    & Text               & {\ul 63.52} & 63.00       & 61.45  & 63.13          & \textbf{65.03}       & 55.60      & 55.66       & {\ul 57.20}           & 56.20        & \textbf{57.58}       & \textbf{18.40} & 14.67          & 12.31          & 15.46          & {\ul 17.47}    \\
\rowcolor[HTML]{EFEFEF} 
\cellcolor[HTML]{FFFFFF}                                    & \textit{Avg.}$\uparrow$               & 38.26       & 41.76       & 37.66  & {\ul 42.87}          & \textbf{45.26}       & 59.97      & 60.85 & \textbf{62.02}  & 61.23        &  {\ul 61.72} & \textbf{18.63} & 16.97          & {\ul 18.40}    & 16.52          & 18.14          \\
\rowcolor[HTML]{EFEFEF} 
\multirow{-4}{*}{\cellcolor[HTML]{FFFFFF}Missing}           & $\Delta_{\textit{Gap}}$$\downarrow$          & 50.53       & 42.48       & 47.59  & {\ul 40.53} & {\textbf{39.54}} & {\ul 8.74} & 10.38       & 9.63            & 10.05        & \textbf{8.28}        & \textbf{0.45}  & 4.59           & 12.17          & 2.11           & {\ul 1.34}     \\
\midrule
\rowcolor[HTML]{FFFFFF} 
\cellcolor[HTML]{FFFFFF}                                    & Image              & 41.92       & 52.92 & 46.50  & \textbf{56.57}       & {\ul 55.58} & 63.64      & {\ul 64.21}       & 63.72           & 61.85        & \textbf{65.78}       & 30.89          & 33.86          & 35.31          & {\ul 35.73}    & \textbf{37.76} \\
\rowcolor[HTML]{FFFFFF} 
\cellcolor[HTML]{FFFFFF}                                    & Text               & 67.28       & {\ul 77.71} & 75.94  & 77.43          & \textbf{78.54}       & 65.09      & 63.66       & \textcolor{gray}{\textbf{67.16*}} & 63.68        & {\ul 65.60}          & 38.09          & \textbf{43.00} & 40.33          & {\ul 42.66}    & 41.80          \\
\rowcolor[HTML]{EFEFEF} 
\cellcolor[HTML]{FFFFFF}                                    & \textit{Avg.}$\uparrow$               & 54.60       & 65.32       & 61.22  & {\ul 67.00} & \textbf{67.06} & 64.37      & 63.94       & {\ul 65.44}     & 62.77        & \textbf{65.69}       & 34.49          & 38.43          & 37.82          & {\ul 39.20}          & \textbf{39.78} \\
\rowcolor[HTML]{EFEFEF} 
\multirow{-4}{*}{\cellcolor[HTML]{FFFFFF}Noisy}             & $\Delta_{\textit{Gap}}$$\downarrow$          & 25.36       & 24.79       & 29.44  & \textbf{20.86}       & {\ul 22.96}          & 1.45       & {\ul 0.55}  & 3.44            & 1.83         & \textbf{0.18}        & 7.20  & 9.14           & {\ul 5.02}           & 6.93           & \textbf{4.04}     \\
\bottomrule

\end{tabular}}
\caption{The experimental result to validate the effectiveness of \textsc{BalGrad} on the UPMC Food-101, Hateful Memes, and MM-IMDb datasets. The best result in each test dataset is boldfaced, and the second best is presented with underlining. ``\textit{Avg.}'' represents the average performance under conditions where one of the modalities is impaired (missing or noisy), while ``$\Delta_{\textit{Gap}}$'' indicates the performance difference. The value that is displayed in \textcolor{gray}{gray*} represents a negative transfer. The unit for ``$\Delta_{\textit{Gap}}$'' is \%p, and the unit for all other values is \%.
}
\label{tab:accents}
\end{table*}

We train with full modality data and evaluate the performance of the VL model under conditions where one modality is entirely impaired across three datasets, as shown in Table~\ref{tab:accents}. ``Full'' refers to the scenario where no modalities are impaired during testing. For the impaired cases (missing and noisy), each modality is impaired according to the specified method. ``\textit{Avg.}'' denotes the average performance when each modality is impaired individually, while ``$\Delta_{\textit{Gap}}$'' represents the performance difference between the image-impaired and text-impaired conditions. A smaller $\Delta_{\textit{Gap}}$ indicates a more balanced model that does not overly rely on a single modality, thereby exhibiting less dominant modality bias.

For the UPMC Food-101 dataset, \textsc{BalGrad} demonstrates the highest performance across all conditions—full, missing image, and missing text. Notably, it improves the performance on the weak modality, text, by 12.5\%p compared to the baseline. Additionally, it achieves the highest average performance and exhibits the smallest gap, effectively mitigating bias despite the dominant influence of the image modality. In the noisy condition, our method shows robustness comparable to AGM~\cite{li2023boosting} and achieves the highest \textit{Avg.} 

\textsc{BalGrad} exhibits the highest performance in conditions where the dominant text modality is missing, as well as in the full modality, \textit{Avg.}, and $\Delta_{\textit{Gap}}$ for the Hateful Memes dataset. OGM-GE~\cite{peng2022balanced} and AGM perform better in the image missing condition than in the full modality condition, indicating a heavy reliance on the text modality, with performance increases of 0.13\%p and 1.56\%p, respectively. In other words, adding the image modality results in a decrease in performance compared to using text alone, exhibiting negative transfer~\cite{wang2019characterizing}. In the noisy condition, \textsc{BalGrad} demonstrates the highest \textit{Avg.} performance and the smallest $\Delta_{\textit{Gap}}$, showcasing that \textsc{BalGrad} sufficiently explores the image modality. 

\begin{figure}[t]
  \includegraphics[width=0.95\linewidth]{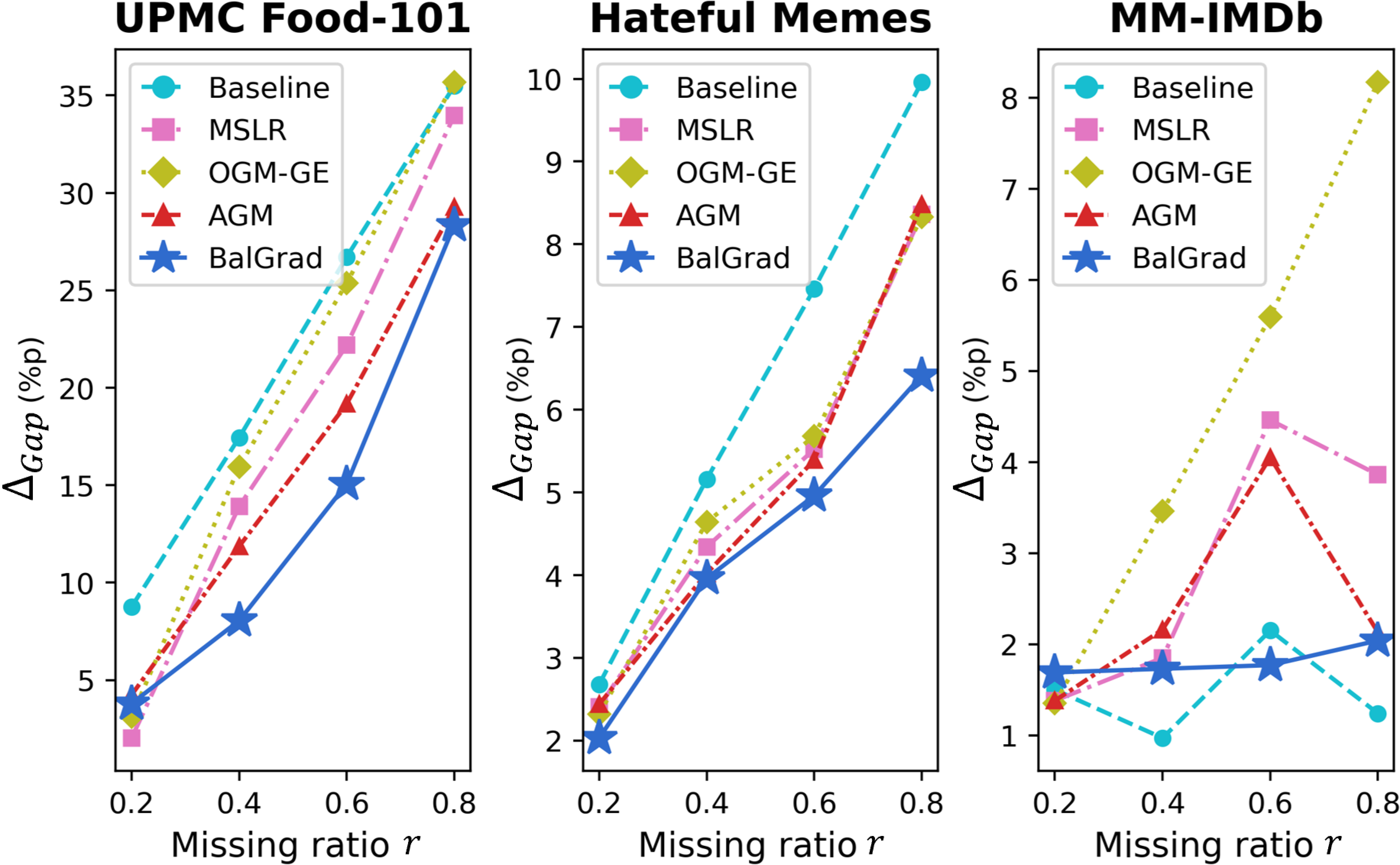}
  \caption{Evaluation on robustness to different missing ratio $r$ of \textsc{BalGrad} and existing methods on UPMC Food-101, Hateful Memes, and MM-IMDb datasets.}
  \label{fig:missing_ratio}
\end{figure}

Furthermore, \textsc{BalGrad} maintains the balance between the modalities even despite the absence of any dominant modality. For the MM-IMDb dataset, our proposed method shows slightly lower performance compared to the baseline but exhibits the second-smallest $\Delta_{\textit{Gap}}$, indicating balanced results without a dominant modality. Although OGM-GE demonstrates high performance, it exhibits a significant imbalance between modalities, as evidenced by the considerably higher gap, which is 10.83\%p more than our method. \textsc{BalGrad} achieves the highest average performance and the lowest gap in the noisy condition, showcasing that our proposed method effectively explores both modalities without being biased towards one.

\begin{table*}[t]
\centering
\resizebox{\textwidth}{!}{%
\addtolength{\tabcolsep}{-0.4em}
\renewcommand{\arraystretch}{1.1}
\begin{tabular}{cc|cccc|cccc|cccc}
\toprule
\rowcolor[HTML]{FFFFFF} 
\multicolumn{2}{c|}{\cellcolor[HTML]{FFFFFF}}                           & \multicolumn{4}{c|}{\cellcolor[HTML]{FFFFFF}\textbf{UPMC Food-101}}                                                                                                                                             & \multicolumn{4}{c|}{\cellcolor[HTML]{FFFFFF}\textbf{Hateful Memes}}                                                                                                                                                                       & \multicolumn{4}{c}{\cellcolor[HTML]{FFFFFF}\textbf{MM-IMDb}}                                                                                                                                                                               \\ \cline{3-14} 
\rowcolor[HTML]{FFFFFF} 
\multicolumn{2}{c|}{\multirow{-2}{*}{\cellcolor[HTML]{FFFFFF}Modality}} & Baseline & \begin{tabular}[c]{@{}c@{}}w/ Gradient\\ reweighting\end{tabular} & \multicolumn{1}{c|}{\cellcolor[HTML]{FFFFFF}\begin{tabular}[c]{@{}c@{}}w/ Gradient\\ projection\end{tabular}} & \textbf{\textsc{BalGrad}} & Baseline   & \cellcolor[HTML]{FFFFFF}\begin{tabular}[c]{@{}c@{}}w/ Gradient\\ reweighting\end{tabular} & \multicolumn{1}{c|}{\cellcolor[HTML]{FFFFFF}\begin{tabular}[c]{@{}c@{}}w/ Gradient\\ projection\end{tabular}} & \textbf{\textsc{BalGrad}} & Baseline    & \cellcolor[HTML]{FFFFFF}\begin{tabular}[c]{@{}c@{}}w/ Gradient\\ reweighting\end{tabular} & \multicolumn{1}{c|}{\cellcolor[HTML]{FFFFFF}\begin{tabular}[c]{@{}c@{}}w/ Gradient\\ projection\end{tabular}} & \textbf{\textsc{BalGrad}} \\ \midrule
\rowcolor[HTML]{FFFFFF} 
\multicolumn{2}{c|}{\cellcolor[HTML]{FFFFFF}Full}                       & 76.01    & {\ul 78.17}                                                       & \multicolumn{1}{c|}{\cellcolor[HTML]{FFFFFF}76.20}                                                            & \textbf{80.32}   & 65.10      & 65.80                                                                                     & \multicolumn{1}{c|}{\cellcolor[HTML]{FFFFFF}{\ul 66.30}}                                                      & \textbf{67.35}   & {\ul 44.09} & \textbf{44.30}                                                                            & \multicolumn{1}{c|}{\cellcolor[HTML]{FFFFFF}42.30}                                                            & 43.19            \\ \midrule
\rowcolor[HTML]{FFFFFF} 
\cellcolor[HTML]{FFFFFF}                                 & Image        & 12.99    & {\ul 22.30}                                                       & \multicolumn{1}{c|}{\cellcolor[HTML]{FFFFFF}19.82}                                                            & \textbf{25.49}   & 64.34      & {\color[HTML]{FFFFFF} \textcolor{gray}{\textbf{66.37*}}}                                       & \multicolumn{1}{c|}{\cellcolor[HTML]{FFFFFF}65.40}                                                            & {\ul 65.86}      & {\ul 18.85} & \textbf{21.48}                                                                            & \multicolumn{1}{c|}{\cellcolor[HTML]{FFFFFF}18.47}                                                            & 18.81            \\
\rowcolor[HTML]{FFFFFF} 
\cellcolor[HTML]{FFFFFF}                                 & Text         & 63.52    & {\ul 64.10}                                                       & \multicolumn{1}{c|}{\cellcolor[HTML]{FFFFFF}63.76}                                                            & \textbf{65.03}   & 55.60      & {\ul 57.03}                                                                               & \multicolumn{1}{c|}{\cellcolor[HTML]{FFFFFF}56.20}                                                            & \textbf{57.48}   & {\ul 18.40} & 17.20                                                                                     & \multicolumn{1}{c|}{\cellcolor[HTML]{FFFFFF}\textbf{18.80}}                                                   & 17.47            \\
\rowcolor[HTML]{EFEFEF} 
\cellcolor[HTML]{FFFFFF}                                 & Avg.↑        & 38.26    & {\ul 43.20}                                                       & \multicolumn{1}{c|}{\cellcolor[HTML]{EFEFEF}41.79}                                                            & \textbf{45.26}   & 59.97      & \textbf{61.70}                                                                            & \multicolumn{1}{c|}{\cellcolor[HTML]{EFEFEF}60.80}                                                            & {\ul 61.67}      & 18.63       & \textbf{19.34}                                                                            & \multicolumn{1}{c|}{\cellcolor[HTML]{EFEFEF}{\ul 18.64}}                                                      & 18.14            \\
\rowcolor[HTML]{EFEFEF} 
\multirow{-4}{*}{\cellcolor[HTML]{FFFFFF}Missing}        & Gap          & 50.53    & {\ul 41.80}                                                       & \multicolumn{1}{c|}{\cellcolor[HTML]{EFEFEF}43.94}                                                            & \textbf{39.54}   & {\ul 8.74} & 9.34                                                                                      & \multicolumn{1}{c|}{\cellcolor[HTML]{EFEFEF}9.20}                                                             & \textbf{8.38}    & {\ul 0.45}  & 4.28                                                                                      & \multicolumn{1}{c|}{\cellcolor[HTML]{EFEFEF}\textbf{0.33}} & 1.34  \\
\bottomrule
\end{tabular}}
\caption{Ablation study results compares performance with and without inter-modality gradient reweighting and inter-task gradient projection to evaluate their impact on modality balance and transfer effects on UPMC Food-101, Hateful Memes, and MM-IMDb datasets. The best results are highlighted in bold, the second-best in italics, and values shown in \textcolor{gray}{gray*} indicate negative transfer. ``$\Delta_{\textit{Gap}}$'' is reported in \%p, while all other values are in \%.}
\label{tab:ablation}
\end{table*}

Additionally, to investigate the robustness under varying degrees of impairment, we mute a specific modality according to the missing ratio $r$, and the results are shown in Figure~\ref{fig:missing_ratio}. For each dataset, we randomly drop a certain percentage $r\%$ of the data from each modality and measure the resulting performance $\Delta_{\textit{Gap}}$. We set missing ratios $r \in \{0.2, 0.4, 0.6, 0.8\}$. Experimental results indicate that \textsc{BalGrad} consistently exhibits a lower gap compared to existing methods across varying missing ratios, demonstrating robustness to impaired modalities. While \textsc{BalGrad} exhibits a slightly larger gap compared to the baseline, it is noteworthy that \textsc{BalGrad} significantly reduces the gap for datasets with dominant modality bias. Additionally, it introduces a small gap for datasets where dominant modality bias is not present.

Additional experimental results on various fusion mechanisms, backbone models, and datasets are provided in Appendix \ref{additional experimental results}. The results demonstrate that \textsc{BalGrad} consistently delivers robust performance across different biases, modality types, datasets, and perturbed conditions, underscoring its effectiveness in synergistically integrating modalities to prevent negative transfer and ensure reliable, real-world multimodal learning.

\subsection{Ablation and Analysis}

\begin{figure}[t]
  \includegraphics[width=\linewidth]{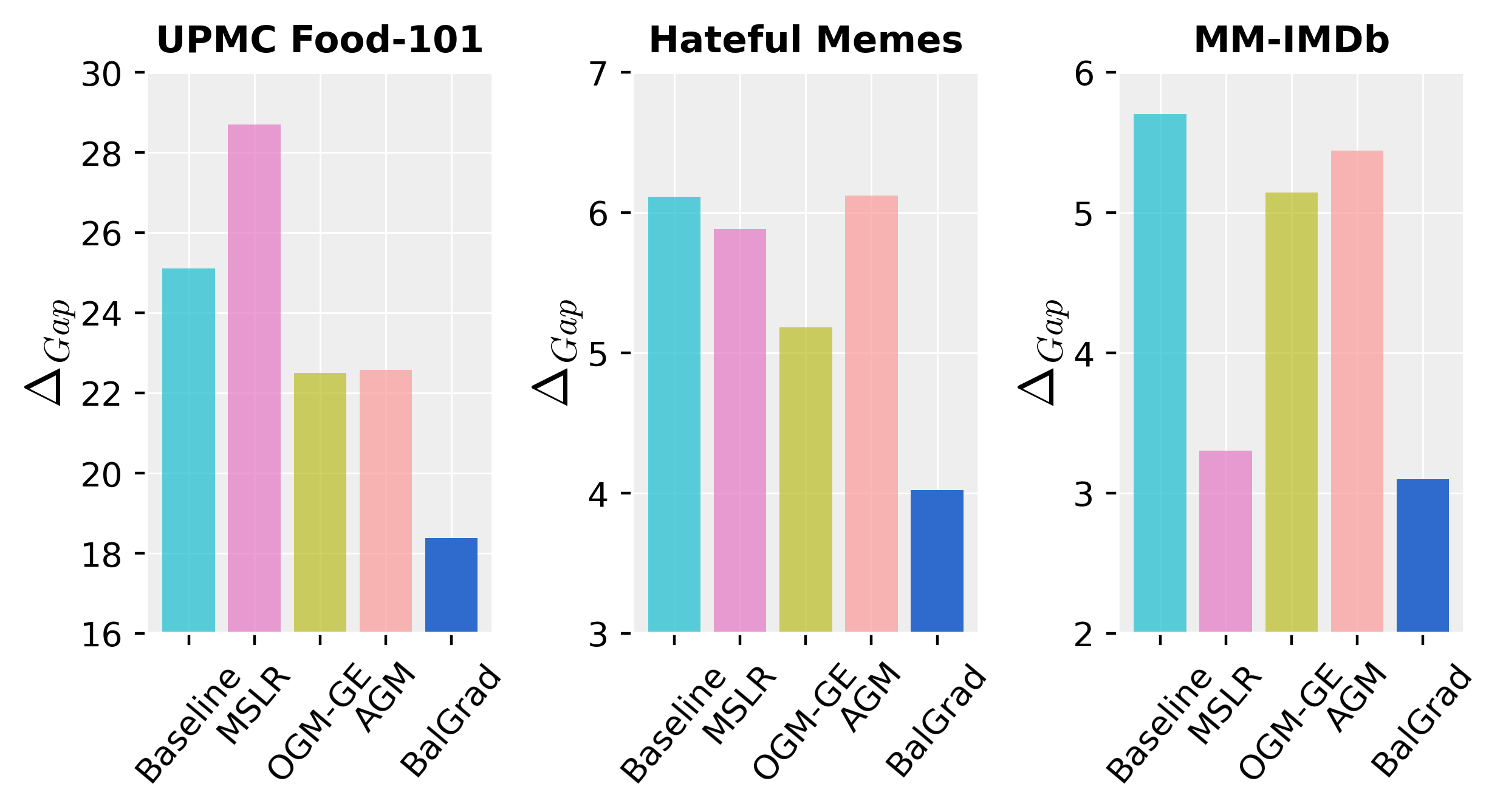}
  \caption{Bar plots comparing the performance of existing methods and \textsc{BalGrad} using BLIP. Each bar represents $\Delta_{\textit{Gap}}$(\%), defined as the performance difference between missing image and missing text conditions.}
  \label{fig:BLIP_gap}
\end{figure}

\begin{figure}[t]
  \includegraphics[width=\linewidth]{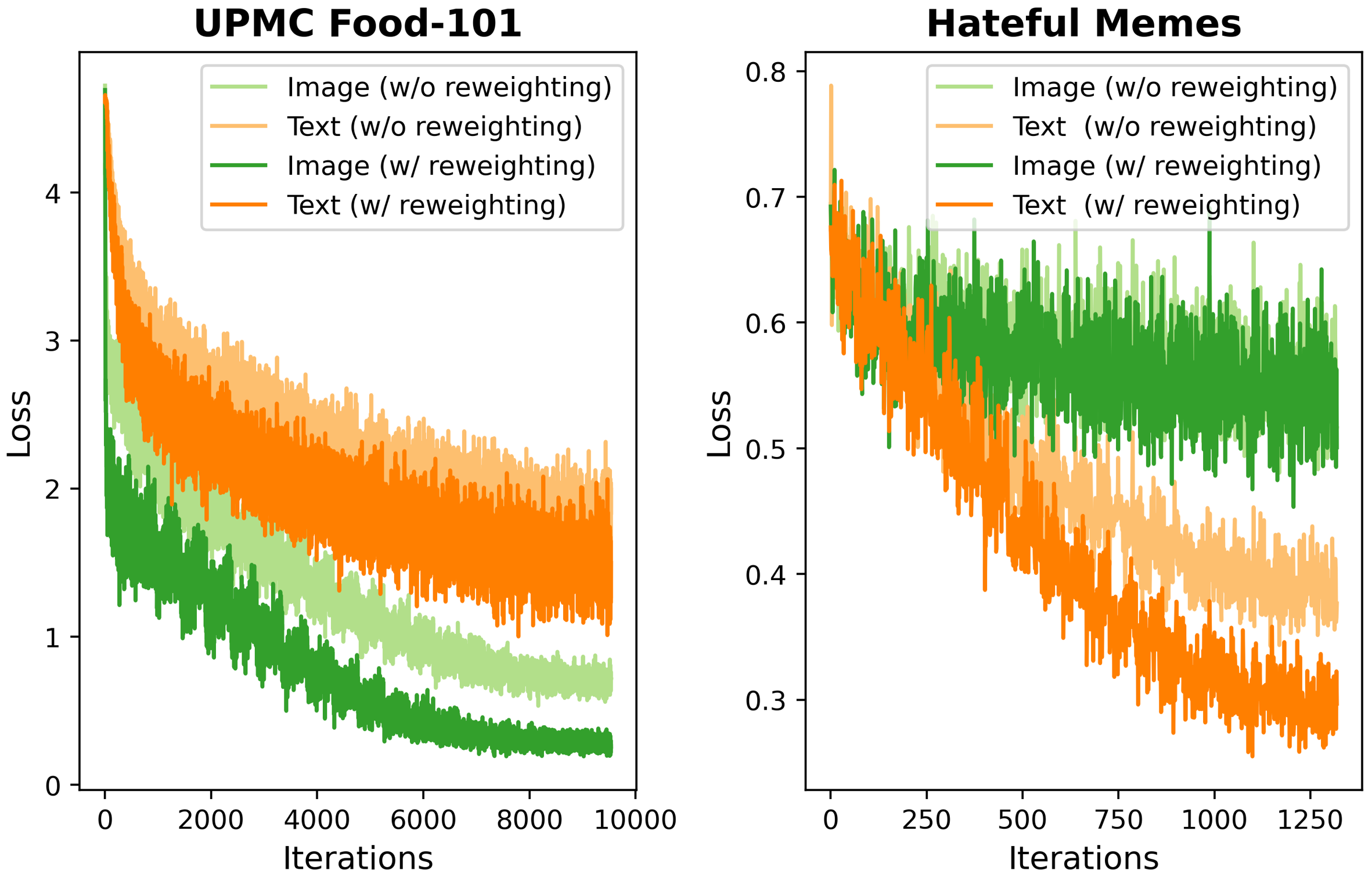}
  \caption{Training iteration loss curves for image and text modalities on the UPMC Food-101 and Hateful Memes datasets, comparing the effects of the existence of inter-modality gradient reweighting.}
  \label{fig:rewighting}
\end{figure}

\noindent \textbf{Analysis of Each Component.} We conduct ablation experiments to assess the impact of gradient reweighting and projection as shown in Table~\ref{tab:ablation}. While gradient reweighting shares a common approach with existing methods~\cite{peng2022balanced, li2023boosting}, helps mitigate modality imbalance, it induces negative transfer in the Hateful Memes dataset and leaves the MM-IMDb dataset overly reliant on text. In contrast, incorporating gradient projection eliminates negative transfer and balances modality use. By aligning the gradient of the target loss with the KL loss term, we reduce reliance on any single modality, effectively preventing negative transfer. These points clarify how our approach differs from existing work and address the gaps in empirical validation and mitigation of negative effects.


\noindent \textbf{Evaluation on Text Decoder-based Vision-Language Model.} 
To examine BALGRAD’s effectiveness in text decoder-based architectures, we conduct additional experiments using BLIP~\cite{li2022blip}, which generates textual outputs from visual inputs via a text decoder. This setup differs from encoder-only VL models and aligns with autoregressive language modeling approaches. As shown in Figure~\ref{fig:BLIP_gap}, BALGRAD achieves the lowest $\Delta_{\textit{Gap}}$ across all datasets, indicating its ability to balance modality contributions in decoder-based VL models. These results highlight BALGRAD’s potential for extension to decoder-only LLMs, as it effectively mitigates dominant modality bias across different VL architectures. 

\noindent \textbf{Ablation on Inter-modality Gradient Reweighting.} To validate the efficacy of inter-modality gradient reweighting, we track the training loss dynamics for each modality on datasets with dominant modality bias (UPMC Food-101 and Hateful Memes), as shown in Figure~\ref{fig:rewighting}. Without reweighting, weights are fixed at $\mathcal{W}^v=1/2$ and $\mathcal{W}^l=1/2$, equally distilling information between modalities. Experimental results show that reweighting leads to faster and more stable convergence of loss for each modality. This supports Proposition 1 in Section~\ref{analysis}, indicating that gradient reweighting optimizes the exploration of individual modalities while maintaining balance in the VL model.

\begin{figure}[t]
  \includegraphics[width=0.95\linewidth]{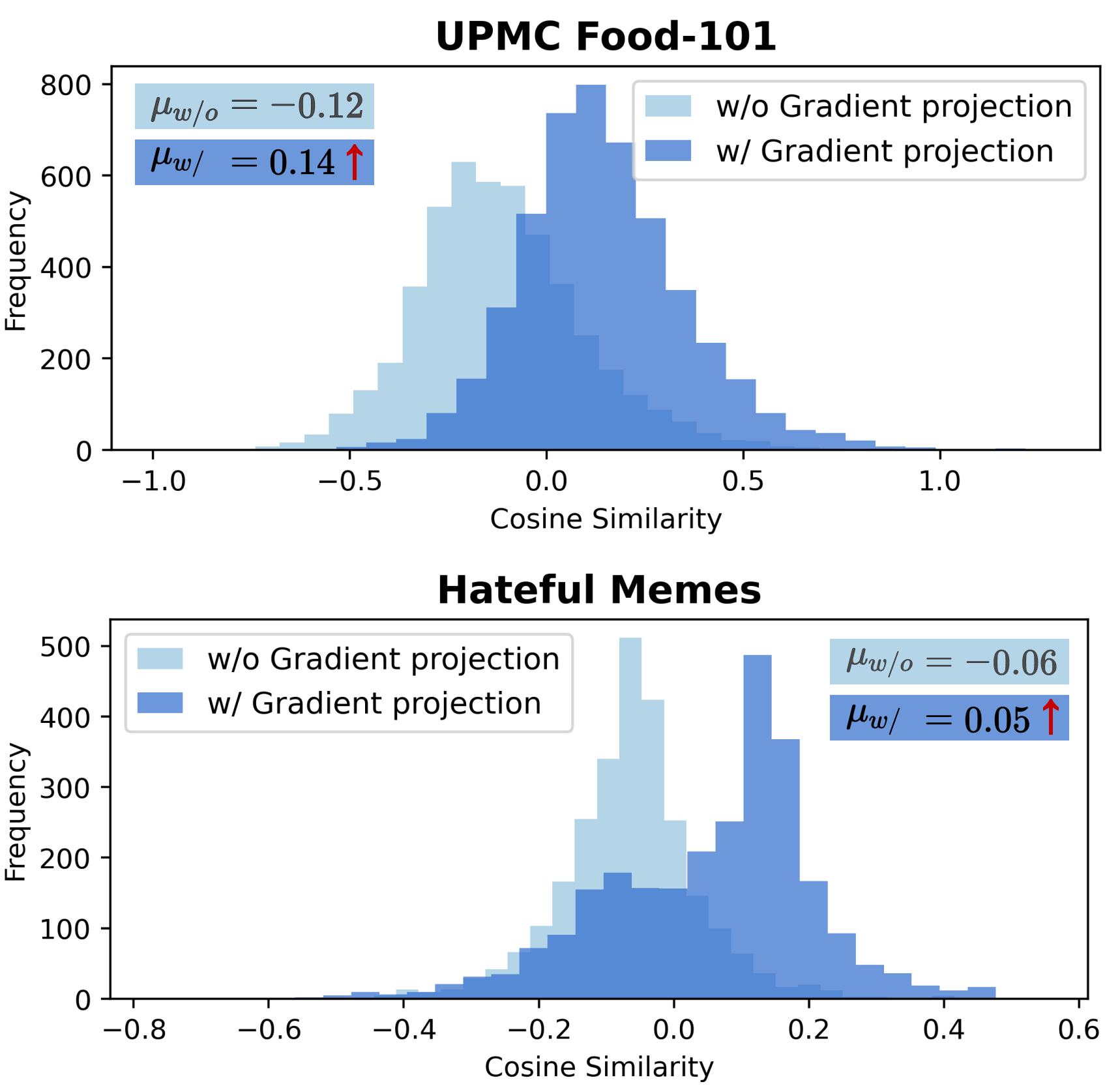}
  \caption{Histogram visualization of the frequency of gradient conflicts between image and text gradients during training iterations on the UPMC Food-101 and Hateful Memes datasets. $\mu_{w/o}$ and $\mu_{w/}$ represent the average cosine similarity values w/o and w/ projection, respectively.}
  \label{fig:hitogram_grad_conflict}
\end{figure}

\noindent \textbf{Analysis on Inter-task Gradient Projection.}To assess the impact of inter-task gradient projection, we visualize the cosine similarity between the gradients of KL divergence ($g_{kl}$) and the target task ($g_{\mathcal{T}}$) throughout the entire training process using histograms, as shown in Figure~\ref{fig:hitogram_grad_conflict}. Without gradient projection, negative similarity between gradients is prevalent throughout training, resulting in imbalanced updates to the target task. Conversely, \textsc{BalGrad}, incorporating inter-task gradient projection, shows a positive mean cosine similarity between gradients, indicating fewer conflicts during training. This suggests that the gradients for the target task are more balanced between the two modalities, leading to more balanced convergence. This reduction in conflicts narrows the performance gap between image and text modalities, mitigating over-reliance on any specific modality, aligning with our analysis in Section~\ref{analysis}.

To further quantify this effect, we conduct an ablation study measuring the frequency of conflicting gradients with and without projection across three datasets, as shown in Table~\ref{tab:projection}. The fraction indicates the percentage of gradient conflicts that occur between the gradients of KL divergence ($g_{kl}$) and the target task ($g_{\mathcal{T}}$) throughout the entire training process. The results demonstrate that there is a high incidence of conflicting gradients across all datasets without projection. In contrast, the use of projection significantly reduces gradient conflicts, especially in datasets with dominant modality bias, such as UPMC Food-101 and Hateful Memes.


\begin{table}[t]
\centering
\resizebox{\linewidth}{!}{%
\addtolength{\tabcolsep}{-0.1em}
\renewcommand{\arraystretch}{1.1}
\begin{tabular}{c|cc|cc|cc}
\toprule
\rowcolor[HTML]{FFFFFF} 
\multirow{2}{*}{}       & \multicolumn{2}{c|}{\textbf{\begin{tabular}[c]{@{}c@{}}UPMC\\ Food-101\end{tabular}}}                            & \multicolumn{2}{c|}{\textbf{\begin{tabular}[c]{@{}c@{}}Hateful\\ Memes\end{tabular}}}                            & \multicolumn{2}{c}{\textbf{MM-IMDb}}                                  \\ \cline{2-7} \rowcolor[HTML]{FFFFFF} 
                        & \multicolumn{1}{c}{Fraction$\downarrow$} & $\Delta_{\textit{Gap}}$$\downarrow$ & \multicolumn{1}{c}{Fraction$\downarrow$} & $\Delta_{\textit{Gap}}$$\downarrow$ & \multicolumn{1}{c}{Fraction$\downarrow$} & $\Delta_{\textit{Gap}}$$\downarrow$ \\
\midrule
\rowcolor[HTML]{FFFFFF} 
w/o Projection& \multicolumn{1}{c}{\large 0.66}           & {\large 43.27}                            & \multicolumn{1}{c}{\large 0.78}           & {\large 10.21}                            & \multicolumn{1}{c}{\large 0.28}           & {\large 4.21}                             \\
\rowcolor[HTML]{EFEFEF} 
w/ Projection  & \multicolumn{1}{c}{\textbf{\large 0.36}}           & \textbf{\large 39.54}                           & \multicolumn{1}{c}{\textbf{\large 0.32}}           & \textbf{\large 8.28}                            & \multicolumn{1}{c}{\textbf{\large 0.26}}           & \textbf{\large 4.04}  \\
\bottomrule
\end{tabular}}
\caption{Ablative results show the fraction of conflicting gradients and $\Delta_{\textit{Gap}}$ on the UPMC Food-101 and Hateful Memes datasets, comparing scenarios without inter-task gradient projection (``w/o Projection'') and with standard \textsc{BalGrad} (``w/ Projection'').}
\label{tab:projection}
\end{table}

\section{Conclusion} 
In this paper, we addressed the challenge of dominant modality bias, where a VL model disproportionately relies on one modality, undermining the contributions of others. Our analysis shows that unaligned gradients and differences in gradient \textit{magnitudes} hinder balanced loss convergence. Based on these findings, \textsc{BalGrad} mitigates this bias by incorporating inter-modality gradient reweighting, which adjusts the KL divergence gradient based on each modality's contribution, and inter-task gradient projection to align task \textit{directions} non-conflictingly. Experiments on UPMC Food-101, Hateful Memes, and MM-IMDb datasets demonstrate that \textsc{BalGrad} effectively reduces dominant modality bias, enhances model robustness, and improves accuracy. These results highlight the potential for more stable and balanced training in VL models, paving the way for future advancements.

\section*{Limitation}
While \textsc{BalGrad} has shown efficacy in mitigating dominant modality bias in VL models, extending this approach to multimodal models with more than two modalities presents additional challenges. When dealing with three or more modalities, the training cost rapidly increases due to the need to consider the relationships between the gradients of each pair of modalities. This increased complexity in gradient management makes the balancing process more computationally intensive and difficult to maintain effectively. Thus, while \textsc{BalGrad} is effective in bi-modal settings, its application in multimodal scenarios requires further refinement to manage the higher computational demands and ensure balanced performance across all modalities.

\section*{Acknowledgment}
This research was supported by the Institute for Information \& Communications Technology Planning \& Evaluation (IITP) grant funded by the Korea government (MSIT) (No. 2021-0-01341, Artificial Intelligence Graduate School Program, Chung-Ang University). This research was also supported by the MSIT (Ministry of Science and ICT), Korea , under the Graduate School of Metaverse Convergence support program (IITP-2025-RS-2024-00418847) supervised by the IITP (Institute for Information \& Communications Technology Planning \& Evaluation).


\bibliography{custom}

\appendix

\section{Appendix of Propositions}

\subsection{Proof of Proposition 1}
\label{proof1}
\noindent \textbf{Proposition 1. (Gradient Effect on Change of Loss)} 
Let the parameters $\theta_v, \theta_l,$ and $\theta_{\mathcal{T}}$ of a multimodal model be updated with gradients $g_v, g_l,$ and $g_{\mathcal{T}}$ using a sufficiently small step size $\lambda>0$, resulting in updated parameters $\hat{\theta}_v, \hat{\theta}_l,$ and $\hat{\theta}_{\mathcal{T}}$. Then the change in the loss function satisfies
\begin{equation}
\begin{aligned}
    \triangle\mathcal{L}
    \;&=\;-2 \,\lambda \,\bigl(g_{\mathcal{T}}^v \cdot g_{\mathcal{T}}^l\bigr)
    \\&
    -\,\lambda \sum_{\,i\in\{v,l,\mathcal{T}\}}\!
      \Bigl(\,g_i \cdot g_i \;+\; g_{\mathcal{T}}^i \cdot g_{\mathcal{T}}^i\Bigr)
    \;+\;
    O(\lambda^2),
\end{aligned}
\end{equation}
where the cross term $-2\,\lambda\,(g_{\mathcal{T}}^v)\cdot(g_{\mathcal{T}}^l)$ captures the interaction between the visual and language gradients and the magnitudes and directions of each gradient $g_{\mathcal{T}}^v$ and $g_{\mathcal{T}}^l$ governs how much the overall loss is reduced.\\

\noindent \textbf{Proof of Proposition 1.} 
\begin{proof}
Let the $\theta_v, \theta_l, \theta_{\mathcal{T}}$ be updated in the direction of negative gradients $g_v, g_l, g_{\mathcal{T}}$ with step size $\lambda>0$.
Then the updated $\hat{\theta}_v, \hat{\theta}_l, \hat{\theta}_{\mathcal{T}}$ are $\theta_v-\lambda g_v, \theta_l-\lambda g_l, \theta_{\mathcal{T}}-\lambda g_{\mathcal{T}}$.
In that case, the change in the loss function with updated parameters is
\begin{align*}
    \triangle\mathcal{L}&=\mathcal{L}(\hat{\theta_{v}}, \hat{\theta}_l, \hat{\theta}_{\mathcal{T}})-\mathcal{L}(\theta_v, \theta_l, \theta_\mathcal{T})
\end{align*}
By the first-order taylor expansion with a point $(\theta_v, \theta_l, \theta_{\mathcal{T}})$,
\begin{equation*}
    \begin{aligned}
    &\mathcal{L}(\hat{\theta_{v}}, \hat{\theta}_l, \hat{\theta}_{\mathcal{T}})-\mathcal{L}(\theta_v, \theta_l, \theta_\mathcal{T})\\&=\mathcal{L}(\theta_v-\lambda g_v, \theta_l-\lambda g_l, \theta_{\mathcal{T}}-\lambda g_{\mathcal{T}})\\&\quad-\mathcal{L}(\theta_v, \theta_l, \theta_\mathcal{T})\\&=\mathcal{L}(\theta_v, \theta_l, \theta_\mathcal{T})\\&\quad+(\theta_v-\lambda g_v-\theta_v)^{T}g_v(\theta_l-\lambda g_l-\theta_l)^{T}g_l\\&\quad+(\theta_{\mathcal{T}}-\lambda g_{\mathcal{T}}-\theta_{\mathcal{T}})^{T}g_{\mathcal{T}}\\&\quad-\mathcal{L}(\theta_v, \theta_l, \theta_\mathcal{T}) ++O(\lambda^2)\\
    &=-\lambda (g_v^T\cdot g_v+g_l^T\cdot g_l)\\&\quad-\lambda(g_{\mathcal{T}}^l+g_{\mathcal{T}}^v)^{T}(g_{\mathcal{T}}^l+g_{\mathcal{T}}^v)\\&\quad +O(\lambda^2) \\&=-2 \lambda g_{\mathcal{T}}^v\cdot g_\mathcal{T}^l-\lambda \sum_{i\in\{v,l,\mathcal{T}\}}(g_i\cdot g_i+g_\mathcal{T}^i \cdot g_\mathcal{T}^i)\\ &\quad+O(\lambda^2)\\
\end{aligned}
\end{equation*}
\end{proof}

\paragraph{Influence of Fusion Methods.}
The term $\bigl(g_{\mathcal{T}}^v\bigr)^\top \bigl(g_{\mathcal{T}}^l\bigr)$ captures how the visual and language gradients interact within the classifier parameters $\theta_{\mathcal{T}}$. 
Different fusion methods yield different dependencies of $g_{\mathcal{T}}^v$ and $g_{\mathcal{T}}^l$ on $v$ and $l$:
\vspace{-2mm}
\begin{itemize}
\item \textbf{Addition:} 
  The fused input is $x = v + l$. Because $v$ and $l$ are merged by simple addition, their representations feed directly 
  into the same part of the classifier. Consequently, $\bigl(g_{\mathcal{T}}^v\bigr)^\top \bigl(g_{\mathcal{T}}^l\bigr)$ often remains significant due to the shared pathway. \vspace{-1mm}
\item \textbf{Concatenation:}
  The fused input is $x = [v; l]$. Each modality is placed in distinct segments of the classifier’s input vector, reducing direct interactions. As a result, $g_{\mathcal{T}}^v$ and $g_{\mathcal{T}}^l$ may be more independent, potentially lowering the cross term  $\bigl(g_{\mathcal{T}}^v\bigr)^\top \bigl(g_{\mathcal{T}}^l\bigr)$. \vspace{-1mm}
\item \textbf{Attention:} The fused input is $x = \mathrm{Attention}(v, l)$. This method can create strong interdependence between $v$ and $l$ within   $\theta_{\mathcal{T}}$. Hence, it $\bigl(g_{\mathcal{T}}^v\bigr)^\top \bigl(g_{\mathcal{T}}^l\bigr)$ can become highly influential since changes $v$ affect $l$ and vice versa 
  through the attention mechanism. \vspace{-1mm}
\end{itemize} \vspace{-2mm}
\noindent Hence, the sign and magnitude of the cross term reflect how strongly the parameters for the 
two modalities are tied together under each fusion strategy.

\subsection{Proof of proposition 2}
\label{proof2}
\noindent\textbf{Proposition 2. (Gradient Conflicts on Loss Reduction with KL Loss)}
Let $\mathcal{G}^\tau = \{\,g_v,\,g_l,\,g_\tau\}$ and $\mathcal{G}^{kl} = \{\,g_v^{kl},\,g_l^{kl},\,0\}$ be the gradients from a target loss $\mathcal{L}_\tau$ and a KL loss $\mathcal{L}_{kl}$, respectively, with parameters $\theta = [\theta_v,\theta_l,\theta_{\tau}]^\top$. Assume the parameters are updated by gradient descent with a small step size $\lambda>0$: $\theta_v' = \theta_v - \lambda\,\bigl(g_v + g_v^{kl}\bigr),\quad
\theta_l' = \theta_l - \lambda\,\bigl(g_l + g_l^{kl}\bigr),\quad
\theta_{\tau}' = \theta_{\tau} - \lambda\,g_{\tau}$.
Then, for the combined loss $\mathcal{L} = \mathcal{L}_\tau + \mathcal{L}_{kl}$, the change in the loss is
\begin{equation}
\begin{aligned}
    \Delta \mathcal{L} &=\mathcal{L}\bigl(\theta'\bigr) -\mathcal{L}\bigl(\theta\bigr) \\&=-\,\lambda \Bigl(\|\mathcal{G}^\tau\|^2 +\,\|\mathcal{G}^{kl}\|^2 +\,2\,(\mathcal{G}^\tau)^\top \mathcal{G}^{kl}\Bigr)\\&+ O(\lambda^2).
\end{aligned}
\end{equation}
In particular, if $(\mathcal{G}^\tau)^\top \mathcal{G}^{kl} < 0,$ the gradients from the target and KL losses \emph{conflict}, reducing the effective loss reduction.\\
\noindent \textbf{Proof of Proposition 2.} 
\begin{proof}
Because $\mathcal{L} = \mathcal{L}_\tau + \mathcal{L}_{kl}$, its gradient is
\begin{equation*}
    \nabla_\theta \mathcal{L} = \mathcal{G}^\tau + \mathcal{G}^{kl}.
\end{equation*}
Under a small step size $\lambda$, a first-order Taylor expansion about $\theta$ gives $\Delta \mathcal{L}\approx-\lambda \,\|\mathcal{G}^\tau + \mathcal{G}^{kl}\|^2$. Since $\mathcal{G}^\tau = \{g_v^\tau, g_l^\tau, g_\tau^\tau\}$ and $\mathcal{G}^{kl} = \{g_v^{kl}, g_l^{kl}, 0\}$, the relevant parameters are updated as: 
\begin{equation*}
\begin{aligned}
&\theta_v' = \theta_v - \lambda\,(g_v^\tau + g_v^{kl})
\\&\theta_l' = \theta_l - \lambda\,(g_l^\tau + g_l^{kl})
\\&\theta_\tau' = \theta_\tau - \lambda\,g_\tau^\tau.
\end{aligned}
\end{equation*}
By decomposing norm:
\begin{equation*}
\begin{aligned}
    \Delta \mathcal{L}=&-\lambda\Bigl(\|g_v^\tau\|^2 + \|g_v^{kl}\|^2 + 2\,g_v^\tau)^\top(g_v^{kl})\\&+\|g_l^\tau\|^2 + \|g_l^{kl}\|^2 + 2\,g_l^\tau)^\top(g_l^{kl})+\|g_\tau^\tau\|^2\Bigr)\\&+ O(\lambda^2)
\end{aligned}
\end{equation*}
Hence, if either $(g_v^\tau)^\top(g_v^{kl}) < 0$ \\
or $(g_l^\tau)^\top(g_l^{kl}) < 0,$ the negative cross-term reduces the effective loss decrease for that modality. 
\end{proof}
\label{sec:appendix}

\section{Further Implementation Details} 
\label{implementation details}

\subsection{Dataset and Evaluation Metrics} 
\noindent \textbf{UPMC Food-101}~\cite{wang2015recipe} is a food classification dataset with 101 categories and 90,840 image-text pairs, involving the classification of food items using both images and textual recipe descriptions; to create a validation split, we extracted 5,000 samples from the training set~\cite{kiela2019supervised}, as the dataset only provides training and testing sets. 

\noindent \textbf{Hateful Memes}~\cite{kiela2020hateful} is designed to detect hate speech by combining image and text modalities, comprising 8,500 training samples, 1,000 validation samples, and 500 test samples. 

\noindent \textbf{MM-IMDb}~\cite{arevalo2017gated} is a multi-label movie genre classification dataset that incorporates poster images and plot descriptions, containing 23 genre tags with 15,552 training samples, 2,608 validation samples, and 7,799 test samples. 

We utilize classification accuracy, AUROC, and F1-Macro as evaluation metrics for the UPMC Food-101, Hateful Memes, and MM-IMDb datasets, respectively.

\subsection{Architecture and Training Scheme} 
In all comparative experiments, we employ ViT~\cite{dosovitskiy2020image} and BERT~\cite{devlin2018bert} as image and text encoders, respectively. We adopt a late concatenation architecture where the embeddings from each modality are concatenated to make the final prediction. We employ linear probing as our fine-tuning strategy, which freezes all the encoder parameters and trains only the embedding and classifier layers. 

We adopt this modular architecture and fine-tuning scheme for several key reasons: First, the modular design of \textsc{BalGrad} allows it to extend to various encoders, easily accommodating different architectures. This flexibility is crucial in real-world scenarios where resources are often constrained. Our structure supports a range of scalable encoder configurations, ensuring adaptability to different resource availability and application requirements. Additionally, in some cases, data access is restricted due to privacy concerns, necessitates the use of pre-extracted features~\cite{cheplygina2019not, kruk2019integrating, menini2020multimodal}, making the application of early fusion-based large VLMs~\cite{liu2024visual} impractical. Also, to focus improvements on \textsc{BalGrad}’s gradient reweighting and projection, we adopt linear probing as a fine-tuning strategy, ensuring that the gains were not merely due to the encoders’ inherent capabilities but to our method’s effectiveness.

As a baseline, we adopt a standard linear probing approach and compare our proposed method against existing methods designed to balance modalities in VL models, specifically MSLR~\cite{yao2022modality}, OGM-GE~\cite{peng2022balanced}, and AGM~\cite{li2023boosting}.

\subsection{Implementation Details} 
We use \texttt{vit-base} and \texttt{bert-base-uncased} checkpoints as the image and text encoders, respectively, loading them from Transformers library~\cite{wolf2020transformers}. The embeddings extracted from each encoder have a dimensionality of 768, and we concatenate these embeddings to form a 1568-dimensional vector, which is then passed to a final classifier. We resize all images to $224 \times 224$ and apply a random horizontal flip for augmentation. For text, the maximum sequence lengths are set to 1024 for MM-IMDb, 512 for UPMC Food-101, and 128 for Hateful Memes. We use the Adam optimizer with a momentum of 0.9 for all experiments, training for 20 epochs with a batch size of 128.

\section{Additional Experimental Results} 
\label{additional experimental results}

\vspace{-2pt}

\begin{figure}[t]
  \includegraphics[width=\linewidth]{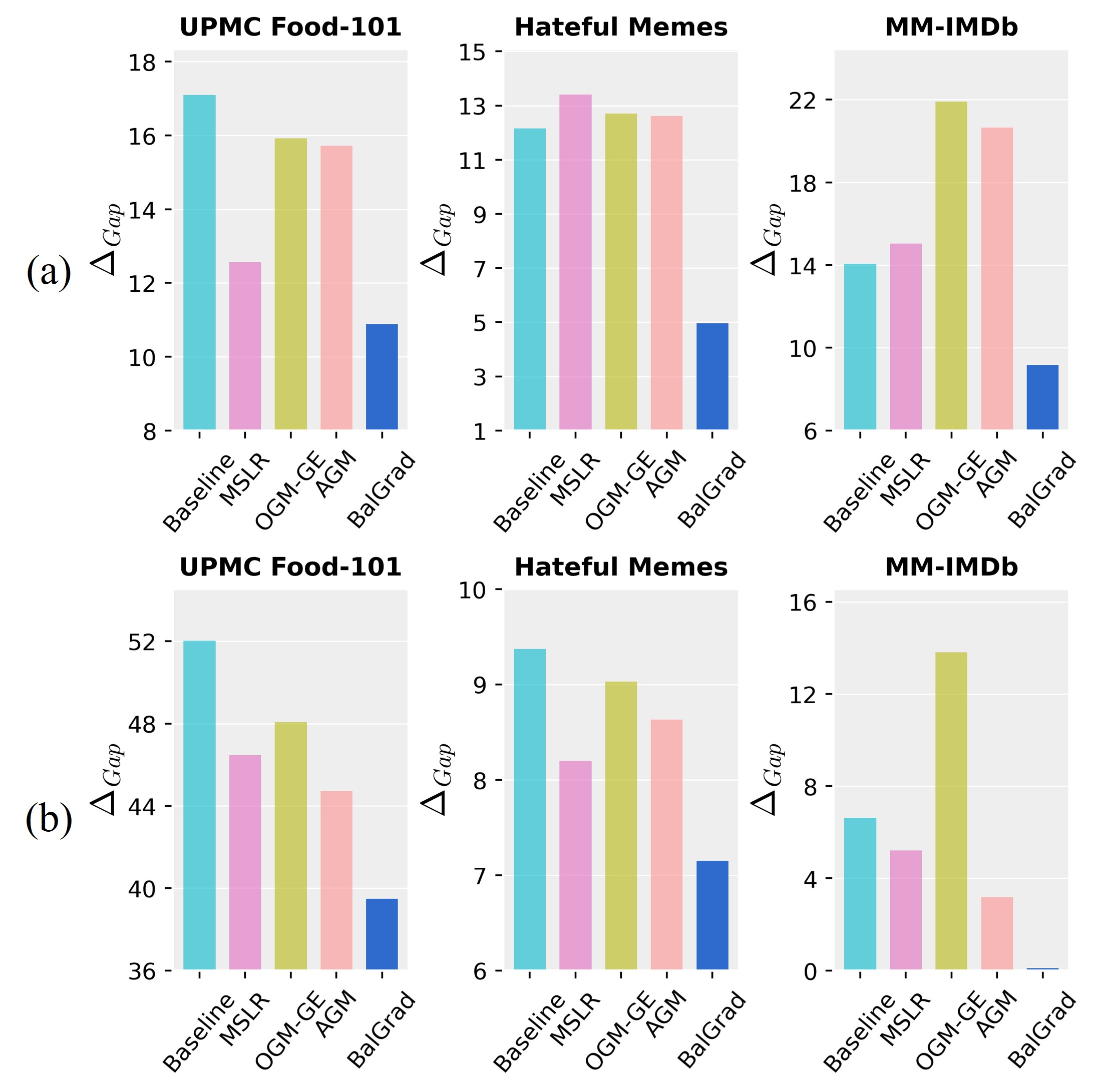}
  \caption{Bar plots illustrating the performance of existing methods and \textsc{BalGrad} with different fusion mechanisms: \textbf{(a)} addition and \textbf{(b)} attention, evaluated on the UPMC Food-101, Hateful Memes, and MM-IMDb datasets. Each bar indicates $\Delta_{\textit{Gap}}$(\%), which quantifies the performance variation between missing image and missing text conditions.}
  \label{fig:gap_addition}
\end{figure}

\begin{figure}[t]
  \includegraphics[width=\linewidth]{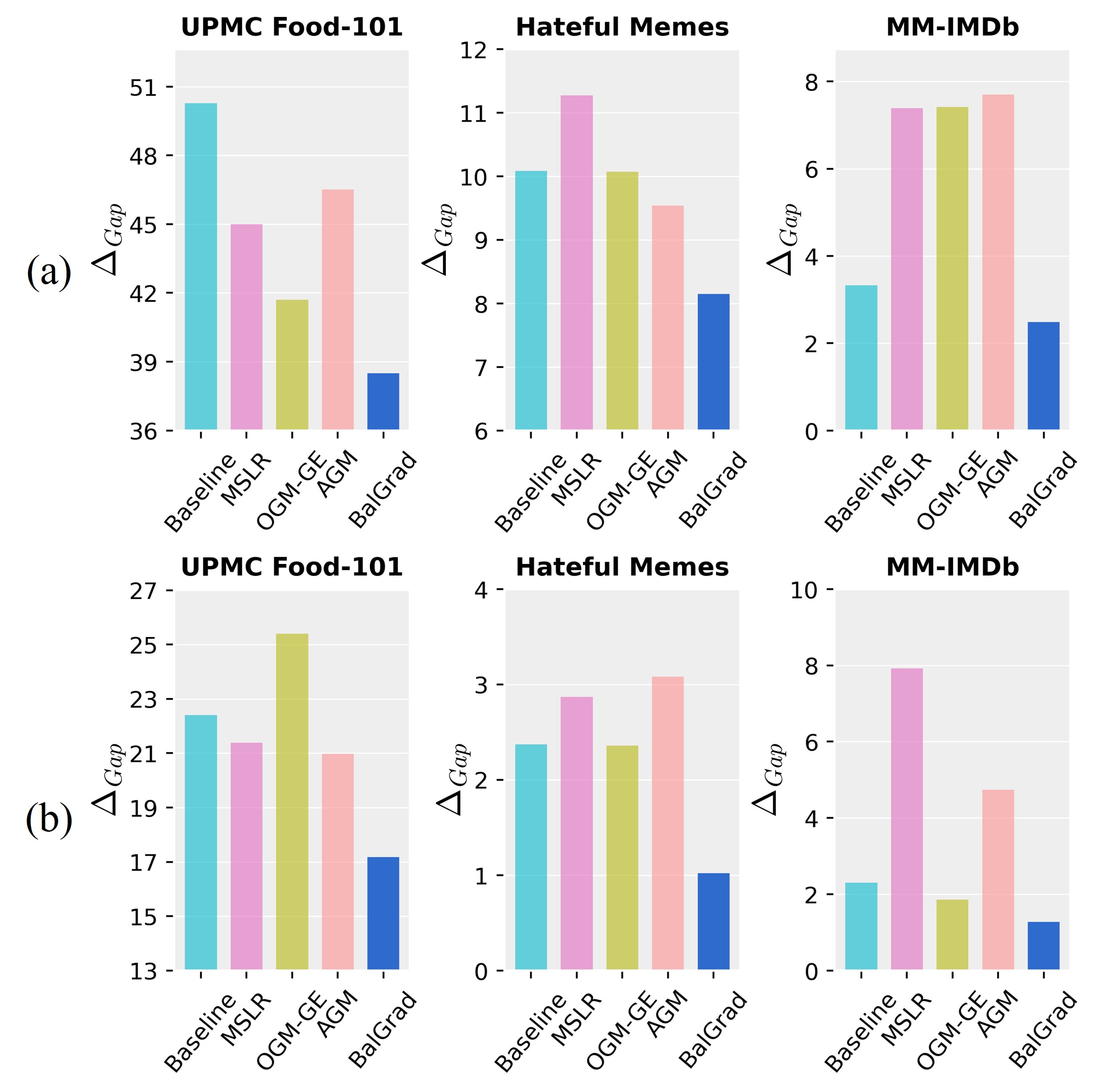}
  \caption{Bar plots presenting the performance comparison between existing methods and \textsc{BalGrad} across different backbone models: \textbf{(a)} ResNet and DistilBERT, and \textbf{(b)} CLIP, on the UPMC Food-101, Hateful Memes, and MM-IMDb datasets. Each bar represents $\Delta_{\textit{Gap}}$(\%), measuring the performance discrepancy under missing image and missing text conditions.}
  \label{fig:gap_resnet-distillbert}
\end{figure}

\subsection{Experimental Results on Different Fusion Mechanisms}

The way embeddings from different modalities are fused can significantly impact a model's ability to capture and leverage cross-modal interactions. We conducted experiments on different fusion strategies in the baseline and \textsc{BalGrad}, specifically exploring element-wise addition and attention-based fusion mechanisms following previous work~\cite{kumar2022hate}. We tested these mechanisms on the UPMC Food-101, Hateful Memes, and MM-IMDb datasets, evaluating the $\Delta_{\textit{Gap}}$ in performance under conditions where either the image or text modality was missing. Results for addition and attention are presented in Figure~\ref{fig:gap_addition}. Across all datasets, \textsc{BalGrad} demonstrated the smallest $\Delta_{\textit{Gap}}$ with both fusion mechanisms, effectively mitigating dominant modality bias. This confirms that \textsc{BalGrad} effectively captures and leverages cross-modal interactions across different fusion mechanisms.

\subsection{Experimental Results on Different Backbone Models}

We conduct extensive experiments across diverse backbone models, underscoring its consistent performance and adaptability to varying architectures and computational resources. Specifically, we employ lower-capacity models—ResNet-50~\cite{he2016deep} for the image encoder and DistilBERT~\cite{sanh2019distilbert} for the text encoder—to assess robustness concerning model size. Additionally, we leverage the widely-used multimodal pretrained VLM, CLIP~\cite{radford2021learning}, for further evaluation due to its strong ability to seamlessly integrate visual and textual information, providing a rigorous test for \textsc{BalGrad}. Experiments were carried out on the UPMC Food-101, Hateful Memes, and MM-IMDb datasets, assessing the performance gap under conditions where either the image or text modality was missing. Results for ResNet-DistilBERT and CLIP are presented in Figure~\ref{fig:gap_resnet-distillbert}. Across all datasets, \textsc{BalGrad} consistently exhibited the smallest $\Delta_{\textit{Gap}}$, effectively balancing the contributions between modalities. 

Intriguingly, while earlier experiments using ViT and BERT encoders on the MM-IMDb dataset showed no over-reliance on a specific modality, our additional studies reveal that conventional methods tend to heavily rely on the text modality, when employing ResNet and DistilBERT. These findings indicate that such bias is influenced not only by the task but also by the choice of backbone model. Our comprehensive experiments affirm that \textsc{BalGrad} effectively mitigates bias irrespective of the backbone model employed, showcasing its superior scalability.

\subsection{Experimental Results with Additional Datasets}

To validate the generalizability of \textsc{BalGrad}, we conduct experiments on two additional datasets: Memotion~\cite{mishra2023memotion} and CUB-200-2011~\cite{welinder2010caltech}. The Memotion dataset, used for classifying the humor level of meme images based on their descriptions, includes annotations such as ``not funny'', ``funny'', ``very funny'', and ``hilarious''. The CUB-200-2011 dataset is a fine-grained bird classification dataset, requiring the categorization of 200 bird species based on images and descriptions. We evaluate each dataset using weighted F1 score and classification accuracy.

The results for the Memotion dataset, presented in Table~\ref{tab:memotion}, show that when the text modality is missing, performance drops significantly more than when the image modality is missing, indicating a bias toward the text modality. \textsc{BalGrad} not only achieves the highest performance with the image modality alone but also excels in the \textit{Avg.} and $\Delta_{Gap}$ metrics, demonstrating effective modality balance.

\begin{table}[t]
\centering
\resizebox{\linewidth}{!}{
\addtolength{\tabcolsep}{-0.1em}
\renewcommand{\arraystretch}{1.1}
\begin{tabular}{cc|ccccc}
\toprule
\multicolumn{2}{c|}{\multirow{2}{*}{Modality}}  & \multicolumn{5}{c}{\textbf{Memotion}}   \\ \cline{3-7} 
\multicolumn{2}{c|}{}        & \multicolumn{1}{c}{Baseline} & \multicolumn{1}{c}{MSLR}   & \multicolumn{1}{c}{OGM-GE}  & \multicolumn{1}{c}{AGM}    & \textbf{\textsc{BalGrad}} \\ 
\midrule
\multicolumn{2}{c|}{Full}    & \multicolumn{1}{c}{70.55}   & \multicolumn{1}{c}{70.36} & \multicolumn{1}{c}{70.28}  & \multicolumn{1}{c}{\textbf{71.12}} & {\ul 70.77}   \\ \midrule
\multicolumn{1}{c}{\multirow{4}{*}{Missing}} & Image & \multicolumn{1}{c}{58.34}        & \multicolumn{1}{c}{59.24}  & \multicolumn{1}{c}{\textbf{59.66}} & \multicolumn{1}{c}{59.54} & 59.48 \\  
\multicolumn{1}{c}{}                        & Text  & \multicolumn{1}{c}{49.29}    & \multicolumn{1}{c}{51.32} & \multicolumn{1}{c}{50.38}  & \multicolumn{1}{c}{{\ul 51.44}} & \textbf{52.78}   \\ 
\multicolumn{1}{c}{}                        & \cellcolor[HTML]{EFEFEF}\textit{Avg.}$\uparrow$   & \multicolumn{1}{c}{\cellcolor[HTML]{EFEFEF}53.82}    & \multicolumn{1}{c}{\cellcolor[HTML]{EFEFEF}55.28} & \multicolumn{1}{c}{\cellcolor[HTML]{EFEFEF}55.02}  & \multicolumn{1}{c}{\cellcolor[HTML]{EFEFEF}{\ul 55.49}} & \cellcolor[HTML]{EFEFEF}\textbf{56.13}   \\ 
\multicolumn{1}{c}{}                        & \cellcolor[HTML]{EFEFEF}$\Delta_{\textit{Gap}}$$\downarrow$   & \multicolumn{1}{c}{\cellcolor[HTML]{EFEFEF}4.53}    & \multicolumn{1}{c}{\cellcolor[HTML]{EFEFEF}{\ul 3.96}}  & \multicolumn{1}{c}{\cellcolor[HTML]{EFEFEF}4.64}   & \multicolumn{1}{c}{\cellcolor[HTML]{EFEFEF}4.05}  & \cellcolor[HTML]{EFEFEF}\textbf{3.35}    \\ 
\bottomrule
\end{tabular}}
\caption{The experimental result of \textsc{BalGrad} on the Memotion dataset. The best result in each test dataset is boldfaced, and the second best is presented with underlining. ``\textit{Avg.}'' represents the average performance under conditions where one of the modality is missing, while ``$\Delta_{\textit{Gap}}$(\%)'' indicates the performance difference.}
\label{tab:memotion}
\end{table}

\begin{table}[t]
\centering
\resizebox{\linewidth}{!}{
\addtolength{\tabcolsep}{-0.1em}
\renewcommand{\arraystretch}{1.1}
\begin{tabular}{cc|ccccc}
\toprule
\multicolumn{2}{c|}{\multirow{2}{*}{Modality}}  & \multicolumn{5}{c}{\textbf{CUB-200-2011}}   \\ \cline{3-7} 
\multicolumn{2}{c|}{}        & \multicolumn{1}{c}{Baseline} & \multicolumn{1}{c}{MSLR}   & \multicolumn{1}{c}{OGM-GE}  & \multicolumn{1}{c}{AGM}            & \textbf{\textsc{BalGrad}} \\ 
\midrule
\multicolumn{2}{c|}{Full}    & \multicolumn{1}{c}{74.71}    & \multicolumn{1}{c}{72.12}  & \multicolumn{1}{c}{75.15}   & \multicolumn{1}{c}{\textbf{76.28}} & {\ul 75.84}               \\ 
\midrule
\multicolumn{1}{c}{\multirow{4}{*}{Missing}} & Image        & \multicolumn{1}{c}{37.38}  & \multicolumn{1}{c}{40.21}   & \multicolumn{1}{c}{39.49} & \multicolumn{1}{c}{{\ul 41.24}} & \textbf{45.47} \\  
\multicolumn{1}{c}{}                        & Text  & \multicolumn{1}{c}{61.24}    & \multicolumn{1}{c}{60.20} & \multicolumn{1}{c}{59.14}  & \multicolumn{1}{c}{{\ul 61.42}} & \textbf{62.72}   \\ 
\multicolumn{1}{c}{}                        & \cellcolor[HTML]{EFEFEF}\textit{Avg.}$\uparrow$   & \multicolumn{1}{c}{\cellcolor[HTML]{EFEFEF}49.31}    & \multicolumn{1}{c}{\cellcolor[HTML]{EFEFEF}50.21} & \multicolumn{1}{c}{\cellcolor[HTML]{EFEFEF}49.32}  & \multicolumn{1}{c}{\cellcolor[HTML]{EFEFEF}{\ul 51.33}} & \cellcolor[HTML]{EFEFEF}\textbf{54.10}   \\ 
\multicolumn{1}{c}{}                        & \cellcolor[HTML]{EFEFEF}$\Delta_{\textit{Gap}}$$\downarrow$   & \multicolumn{1}{c}{\cellcolor[HTML]{EFEFEF}11.93}    & \multicolumn{1}{c}{\cellcolor[HTML]{EFEFEF}9.99}  & \multicolumn{1}{c}{\cellcolor[HTML]{EFEFEF}{\ul 9.83}}   & \multicolumn{1}{c}{\cellcolor[HTML]{EFEFEF}10.09}  & \cellcolor[HTML]{EFEFEF}\textbf{8.63}    \\ 
\bottomrule
\end{tabular}}
\caption{The results of \textsc{BalGrad} on the CUB-200-2011 dataset are presented. The highest performance in each test dataset is shown in bold, with the second-highest underlined. ``\textit{Avg.}''  reflects the average performance when one modality is absent, and ``$\Delta_{\textit{Gap}}$(\%)'' denotes the performance difference.}
\label{tab:cub}
\end{table}

As shown in Table~\ref{tab:cub}, the CUB-200-2011 dataset exhibits a strong reliance on the image modality. \textsc{BalGrad} outperforms AGM by more than 4\%p in accuracy when the image modality is missing and achieves the smallest $\Delta_{Gap}$ at 8.63\%, demonstrating its superiority in handling fine-grained classification tasks even under challenging conditions.


\end{document}